\documentclass[letterpaper,twocolumn,10pt]{article}
\usepackage{usenix}
\usepackage{hyperref}
\usepackage{tikz}
\usepackage{amsthm}
\newtheorem{theorem}{Theorem}

\newtheorem{definition}{Definition}
\usepackage{amsmath}
\usepackage{algorithm}
\usepackage{algpseudocode}
\usepackage{algcompatible}
\usepackage{amssymb}
\usepackage{verbatim}
\usepackage{amsthm}
\usepackage{makecell}
\usepackage{tabularx}
\usepackage{booktabs}
\usepackage{hyperref}  
\usepackage{xurl}
\usepackage{algorithm}
\usepackage{algpseudocode}
\usepackage{float}
\usepackage{graphicx}   %
\usepackage{multirow}
\usepackage{subcaption} %
\usepackage{enumitem}

\usepackage[T1]{fontenc}
\usepackage{microtype}
\microtypecontext{spacing=nonfrench}
\microtypesetup{protrusion=true, expansion=true}
\UseMicrotypeSet[protrusion]{basicmath}
\usepackage{xspace}
\usepackage[available]{usenixbadges}
\usepackage{fancyhdr}
\newif\ifAnon
\Anonfalse
\newif\ifUsenix
\newif\ifArxiv

\Arxivtrue
\Usenixfalse

\newif\ifdraft
\draftfalse
\newif\ifNotes
\Notestrue

\ifdraft
\else
\Notesfalse
\fi

\renewcommand{\paragraph}[1]{
  \smallskip\noindent\textbf{#1}
}

\ifNotes
    \newcommand{\colorcomment}[2]{\leavevmode\unskip\space{\color{#1}#2}\xspace}
    
\else
    \newcommand{\colorcomment}[2]{\leavevmode\unskip\relax}
    
\fi

\definecolor{darkgreen}{rgb}{0,0.65,0}
\definecolor{darkviolet}{HTML}{9400D3}
\definecolor{wildwatermelon}{HTML}{FF43A4}
\definecolor{bluegreen}{HTML}{2ed285}
\ifdraft
    \newcommand{\taggedcolorcomment}[3]{\colorcomment{#1}{[\textbf{#2}: #3]}}
    \newcommand{\todo}[1]{\colorcomment{red}{[Todo: #1]}}
    
    \newcommand{\refs}[1]{\colorcomment{red}{[#1]}}
    \newcommand{\crap}[1]{\colorcomment{red}{[crap: #1]}}
\else
    \newcommand{\taggedcolorcomment}[3]{}
    \newcommand{\todo}[1]{}
    
    \newcommand{\refs}[1]{}
    \newcommand{\crap}[1]{}
\fi

\newcommand{\secref}[1]{\S\ref{#1}\xspace}
\newcommand{\secsref}[2]{\S\ref{#1}--\ref{#2}\xspace}
\newcommand{\figref}[1]{Fig.~\ref{#1}\xspace}

\newcommand{\tabref}[1]{Table~\ref{#1}\xspace}

\newcommand{\eqnref}[1]{Eq.~\ref{#1}\xspace}
\newcommand{\appref}[1]{App.~\ref{#1}\xspace}
\newcommand{\algoref}[1]{Alg.~\ref{#1}\xspace}

\newcommand{\ml}{ML\xspace}
\newcommand{\defense}{Sy-FAR\xspace}
\newcommand{\specnorm}{SpecNorm\xspace}
\newcommand{\faal}{FAAL\xspace}
\newcommand{\pubfig}{PubFig\xspace}
\newcommand{\pubfigsiblings}{\text{\pubfig}$_{\text{SIB}}$\xspace}
\newcommand{\vgg}{VGG-16\xspace}
\newcommand{\vit}{ViT\xspace}
\newcommand{\pubfigvit}{\text{\pubfig}$_{\text{\vit}}$\xspace}

\newcommand{\Penalty}{\mathrm{penalty}}
\newcommand{\SymLoss}{\mathcal{L}_{\mathrm{sym}}}

\newcommand{\ConfMat}[2][]{%
  \if\relax\detokenize{#1}\relax
    C%
  \else
    C_{#1#2}%
  \fi
}

\begin{document}
\date{}

\title{\Large \bf Sy-FAR: Symmetry-based Fair Adversarial Robustness}

\author{
\ifAnon
    Anonymous Submission
\else
    {\rm Haneen Najjar}\\
    Tel Aviv University
    \and
    {\rm Eyal Ronen}\\
    Tel Aviv University
    \and
    {\rm Mahmood Sharif}\\
    Tel Aviv University
\fi
} %

\maketitle

\begin{abstract}

Security-critical machine-learning (ML) systems, such as face-recognition systems, are susceptible to adversarial examples, including real-world physically realizable attacks.
Various means to boost ML's adversarial robustness have been proposed; however, they typically induce unfair robustness: It is often easier to attack from certain classes (e.g., individuals) or groups (e.g., genders) than from others.
Several techniques have been developed to improve adversarial robustness while seeking perfect fairness between classes. 
Yet, prior work has focused on settings where security and fairness are less critical
(e.g., classifying objects such as cars and ships).

Our insight is that achieving perfect parity in realistic fairness-critical tasks, such as face recognition, is often infeasible---some classes (e.g., siblings) may be highly similar, leading to more misclassifications between them. 
Instead, we suggest that seeking symmetry---i.e., attacks from class $i$ to $j$ would be as successful as from $j$ to $i$---is more tractable.
Intuitively, symmetry is desirable because class resemblance is a symmetric relation in most domains.
Additionally, as we prove theoretically, symmetry between individuals induces symmetry between any set of sub-groups, in contrast to other fairness notions where group-fairness is often elusive.

We develop Sy-FAR, a technique to encourage symmetry while also optimizing adversarial robustness and extensively evaluate it using five datasets, with three model architectures, including against targeted and untargeted realistic attacks.
The results show Sy-FAR significantly improves fair adversarial robustness compared to state-of-the-art methods.
Moreover, we find that Sy-FAR is faster and more consistent across runs.
Notably, Sy-FAR also ameliorates another type of unfairness we discover in this work---target classes that adversarial examples are likely to be classified into become significantly less vulnerable after inducing symmetry.

\end{abstract}

\section{Introduction}

Adversarial examples---perturbed inputs that lead machine-learning (\ml{}) models to misclassify at deployment time---pose a profound challenge to \ml{}~\cite{biggio-2013, szegedy2014intriguing}.
Notably, adversarial examples are not a hypothetical risk to \ml{}, but can be implemented under real-world constraints, thus harming \ml{} systems' integrity in practical settings~\cite{Evtimov17Signs, sharif2016accessorize}.
For instance, adversaries may physically realize attacks against face-recognition systems via eyeglasses they can don to evade surveillance or impersonate others~\cite{sharif2016accessorize, Sharif19AGNs}.

Realizing the imminent risk to \ml{} systems that are becoming increasingly ubiquitous, researchers have explored various directions to enhance models' adversarial robustness (e.g.,~\cite{Cohen19RandSmooth, goodfellow2015advtraining, Katz19Marabou, madry2018towards, Xu18Squeeze}).
Different classes of defenses have been proposed, %
ranging from 
inference-time countermeasures that may refine classification boundaries or derive security guarantees (e.g.,~\cite{Cohen19RandSmooth, Katz19Marabou}) to adversarial training, which augments training with attacks to improve robustness (e.g.,~\cite{goodfellow2015advtraining, madry2018towards, tong2021facesec}). 
From these, adversarial training is particularly appealing due to significant improvements in robust accuracy (i.e., accuracy under attacks) without increasing inference-time latency. %

\begin{figure}[t!]
    \centering
    \begin{minipage}[b]{0.95\linewidth}
        \centering
        \includegraphics[width=0.95\linewidth]{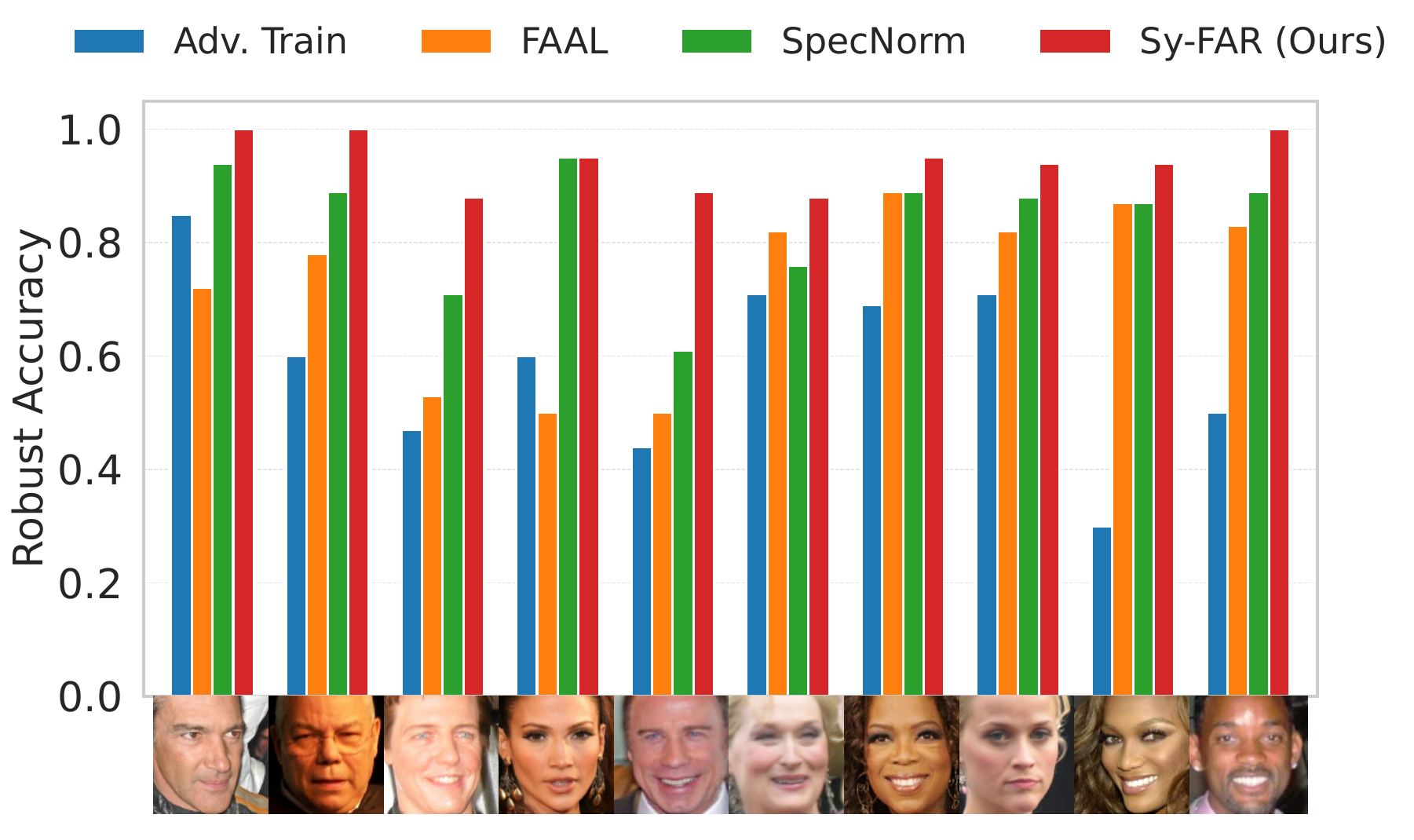}
    \end{minipage}
    \hfill
    \caption{Per-class robust accuracy of face-recognition models trained using different defensive methods on a subset of \pubfig{} dataset~\cite{attribute_classifiers} using the \vgg{} architecture~\cite{simonyan2015very}.
    The models were trained to recognize a set of ten celebrities with an equal number of males and females (\secref{sec:experiment_setup}).
    We report results obtained with four methods: adversarial training~\cite{tong2021facesec}, two leading approaches for enhancing fair source-class adversarial robustness (\faal{}~\cite{zhang2024towards} and \specnorm{}~\cite{jin2025enhancing}), and our proposed method, \defense{}. 
    Adversarial examples were created with (untargeted) eyeglass attacks~\cite{sharif2016accessorize}.
    \vspace{-5pt}
    }
    \label{fig:pubfig_robust_fairness_all}
\end{figure}

\paragraph{Fairness Challenges in Robust Classification.}
While defenses substantially improve adversarial robustness, they suffer from 
fairness issues, due to disparities between classes (e.g., individuals in the case of face recognition) or groups (e.g., genders or ethnicities)~\cite{nanda2021fairness, rosenberg2023fairness, xu2021robust}.
In particular, while defenses enhance adversarial robustness, they often do so unevenly across classes. 
Said differently, it is often significantly easier to produce adversarial examples for inputs from certain (source) classes than inputs from other classes.
Besides ethical implications, these disparities in robust accuracy across classes---a phenomenon termed \emph{unfair source-class adversarial robustness}---have direct security implications.
For example, if such disparities exist in a face-recognition system in surveillance, then certain subjects may still easily evade detection although the overall robustness seems high.
As \figref{fig:pubfig_robust_fairness_all} shows, source-class unfairness is a significant issue in defended face-recognition models: certain subjects receive significantly lower robust accuracy than others (e.g., Antonio Banderas, first from the left, and Beyonc\'e, second from the right, for the adversarially trained model).

Yet another fairness issue facing defenses that, to our knowledge, we characterize for the first time (\secref{sec:res:target}), is that of \emph{unfair target-class robustness}.
Specifically, we find that certain classes are more likely to be the ``sinks'' of misclassified adversarial examples.
Put simply, adversarial examples have higher probability of being misclassified \emph{into} certain classes than others.
This form of unfairness directly affects security: For example, in a face-recognition system that exhibits target-class unfairness, specific individuals would be at a higher risk of impersonation than others.
Hence, in addition to countering unfair source-class adversarial robustness, as has been the focus of prior work (e.g.,~\cite{jin2025enhancing, zhang2024towards}), it is critical to address unfair target-class adversarial robustness.

Last, but not least, besides achieving fairness for individuals, we should also seek to enhance fairness for different groups, such as different genders or ethnicities, so no group remains disadvantaged.
Prior work has demonstrated that unfair source-class robust accuracy is exhibited not only for individual classes, but also for groups~\cite{rosenberg2023fairness, nanda2021fairness}.
A core challenge arising when seeking fairness for groups is that their number may be excessively large (exponential in the number of classes), rendering it prohibitive to improve fairness across \emph{all} groups~\cite{rothblum2018probably}.
Accordingly, to our knowledge, no existing technique is able to promote fair adversarial robustness for an arbitrary group that may be unknown at training time.

\paragraph{Established Solutions.}
Several methods have been proposed to ameliorate unfair source-class robustness~\cite{jin2025enhancing, li2023wat, ma2022fat, sun2023improving, wei2023cfa, xu2021robust, zhang2024towards}. 
These roughly share the same principle: 
While optimizing for overall benign and robust accuracy (so these are preserved or improved compared to standard training), they also seek to directly decrease the gap in robust accuracy between classes (e.g., by assigning higher weights for classes with lower robustness) or to increase the robust accuracy of the class with minimal robustness.
In doing so, leading approaches such as \faal{}~\cite{zhang2024towards} and \specnorm{}~\cite{jin2025enhancing} substantially improve source-class fairness.
However, as shown in \figref{fig:pubfig_robust_fairness_all}, specific classes still remain highly vulnerable even under these defenses (e.g., John Travolta, fifth from the left).
Notably, previous work evaluated setups where fairness and security have little implications, as they focus on artificial object classification tasks (mainly, the CIFAR datasets~\cite{krizhevsky2009learning}) and impractical attacks where adversaries introduce imperceptible perturbations (bounded in $\ell_p$-norm) to induce misclassifications.
As a result, prior work's ability to improve fairness while preserving robustness and benign accuracy (i.e., accuracy on clean inputs) in realistic fairness- and security-critical settings remains unknown. 
We address this gap by porting previous approaches to realistic face-recognition settings (while also evaluating them on the original setups).

\paragraph{Our Solution.}
In this work, we argue that achieving perfect source-class fairness, as sought by prior work~\cite{jin2025enhancing, li2023wat, ma2022fat, sun2023improving, wei2023cfa, xu2021robust, zhang2024towards}, may often be infeasible.
For instance, assume a setting where face recognition is trained to recognize unrelated individuals as well as two siblings (or even twins). 
Naturally, misclassifications between the siblings would be more challenging to prevent than misclassifications between other individuals.
Achieving perfect parity in this setting may even be counterproductive, as it may require reducing the robust accuracy of classes other than the siblings.

As an alternative, we propose to tackle fairness issues in adversarially robust classification via a novel technique---by \emph{encouraging symmetry between classes}, as part of a framework for {\emph Sy}mmetry-based {\emph F}air {\emph A}dversarial {\emph R}obustness ({\emph \defense}).
Besides optimizing for benign and robust accuracy, \defense{} trains models that would be nearly as likely to misclassify adversarial examples from class $i$ to $j$ as they are to misclassify in the other direction %
(\secref{sec:technical_approach}).
Intuitively, as class resemblance is typically a symmetric relation, it is natural to seek symmetry, such that misclassifications (of adversarial examples) between class pairs would be equally likely.
Consequently, no class would be significantly more disadvantaged against the other.
Crucially, we find that promoting symmetry boosts fair adversarial robustness from both source- and target-class perspectives (\secref{sec:experiment_results}).
Furthermore, as we formally prove, ensuring symmetry for individual classes also leads to symmetry between arbitrary groups (\secref{sec:fairness_subgrp}).
Consequently, \defense{}, in a way completely agnostic to group definitions, %
encourages symmetry between arbitrary groups and, as a by-product, enforces fair source- and target-group  robustness for any arbitrary partition of classes into groups.

\paragraph{Our Contributions.}
We make several key contributions:

\begin{itemize}[leftmargin=10pt, itemsep=1pt]

\item We propose the \defense{} (\secref{sec:technical_approach}), a defense encouraging symmetry in misclassification patterns of adversarial examples, while optimizing benign and robust accuracy.

\item We suggest a tractable, computationally efficient method to optimize symmetry (through a differentiable loss function) as part of \defense{}, lending itself to efficient integration in the training of neural networks (\secref{sec:tech:alg}).

\item We extensively evaluate \defense{} on two vision tasks (face and object recognition), using three model architectures, and two attack types (\secref{sec:experiment_setup}). 
Importantly, in contrast to pervious work, we consider a realistic adversary producing adversarial eyeglasses that would lead face-recognition to misclassify~\cite{sharif2016accessorize}. %
Thus, our work considers a real-world setting where adversarial robustness and fairness are critical.
Moreover, we compare \defense{} empirically to \faal{}~\cite{zhang2024towards} and \specnorm{}~\cite{jin2025enhancing}, two state-of-the-art techniques. %

\item We find that, compared to past methods, by improving symmetry, \defense{} more substantially improves fair source-class adversarial robustness (\secref{sec:experiment_results:source}). 
For instance, on challenging face-recognition tasks with siblings where previous methods are particularly asymmetric, \defense{} reduces the gap between the most and least robust classes by $\geq$41\%. compared to baselines.
Importantly, \defense{} also preserves and sometimes even improves benign and robust accuracy.

\item We identify, to our knowledge for the first time, that \ml{} models exhibit unfair target-class adversarial robustness (\secref{sec:experiment_results:target}). 
Additionally, we find that \defense{} helps ameliorate this form of unfairness by encouraging symmetry. For example, on face recognition, \defense{} makes the most vulnerable class $\geq$56\% less likely to be erroneously selected as the output of adversarial examples compared to baseline methods.

\item We theoretically prove that encouraging symmetry for classes leads to symmetry for \emph{any} partitioning of classes into mutually exclusive groups (\secref{sec:fairness_subgrp}). As a result, \defense{} encourages symmetry between any pair of arbitrary groups, thus promoting fair source- and target-group adversarial robustness.

\item Interestingly, we find that \defense{} is more computationally efficient (\secref{sec:experiment_results:time}) and consistent at optimizing models for fairness and robustness (\secref{sec:experiment_results:stability}) than alternative approaches.

\end{itemize}

The paper proceeds as follows.
We next discuss related work (\secref{sec:relwork}), lay out the threat model (\secref{sec:threat}), and present the technical approach behind \defense{} (\secref{sec:technical_approach}).
Subsequently, we describe our experimental setup (\secref{sec:experiment_setup}) before we present our results for fair (class-level) source- and target-class adversarial robustness (\secref{sec:experiment_results}) and group fairness (\secref{sec:fairness_subgrp}).
We wrap the paper with discussion of limitations and future work (\secref{sec:limitations}) and concluding remarks (\secref{sec:conclude}).

\section{Related Work}
\label{sec:relwork}

\subsection{Adversarial Examples}

The proliferation of \ml{} has raised researchers' interest in \ml{}'s implications on system security.
Researchers have shown various attacks on \ml{} privacy (e.g.,~\cite{membership-inference-paper}), integrity (e.g.,~\cite{goodfellow2015advtraining, szegedy2014intriguing, Biggio12Pois}), and availability (e.g.,~\cite{shumailov2021sponge}), among others. 
Specifically, adversarial examples have received special attention due to the potential risk they pose to deployed safety- and security-critical systems during deployment: 
Numerous attacks targeting models of varied modalities under different assumptions and with various attack objectives were proposed since the initial discovery of adversarial examples (e.g.,~\cite{Evtimov17Signs, papernot-2016-blackbox, mahmood-binary, sharif2016accessorize, Schon18AdvSound, zouGCG-UniversalTransferableAdversarial2023}).

While one class of adversarial examples leverages imperceptible adversarial perturbations, bounded in $\ell_p$-norm, to mislead \ml{} models~\cite{carlini2019evaluating, Croce20Auto, goodfellow2015advtraining, Moosavi16DeepFool, szegedy-2014-intriguing},
another class targets more realistic settings where adversaries may be more bound by practical constraints~\cite{Evtimov17Signs, mahmood-binary, Schon18AdvSound, sharif2016accessorize, Sharif19AGNs, Xu16PDF}.
For instance, in the \emph{eyeglass attack} against face recognition systems~\cite{sharif2016accessorize}, adversaries may need to physically realize artifacts that are robust to printing and camera noise to mislead systems under real-world constraints.
Different types of attacks may have untargeted or targeted variants~\cite{jsma-paper}---i.e., aiming to achieve arbitrary misclassification or misclassification into specific classes, respectively---as well as other variants suitable for other assumptions (e.g., white- and black-box~\cite{Brendel18DBA, liu2017delving, papernot-2016-blackbox}).
In this work, we study the fairness properties of \ml{} models subject to different attack types, with a particular focus on the realistic eyeglass attack.

\paragraph{Defenses.}
Different forms of defenses against adversarial examples have been proposed, including, but not limited to, defenses that 
detect attacks (e.g.,~\cite{Metzen17Detector}); 
filter out adversarial perturbations (e.g.,~\cite{Xu18Squeeze});
smoothen the classification boundaries to reduce model vulnerability (e.g.,~\cite{Carlini23Certified, Cohen19RandSmooth, Salman19Robust});
verify robustness against specific adversaries (e.g.,~\cite{Katz19Marabou}); and
adversarially train models to inherently increase their robustness (e.g.,~\cite{kurakin2017adversarial, madry2018towards}).
Due to its intuitive nature, its ability to improve adversarial robustness in a practical manner against different attack types, and absence of impact on model's inference time, %
adversarial training is particularly appealing and was widely studied (e.g.,~\cite{goodfellow2015advtraining, Kurakin16AdvTrain, li2023wat, Lucas23AdvTrain, madry2018towards, Shafahi19PGD, tong2021facesec, Wong20Fast}). 
Accordingly, in this work, we focus on improving the fairness properties of adversarially trained models, especially against more realistic attacks.
As adversarial training is often combined with other defenses (e.g., randomized smoothing~\cite{Salman19Robust}), we expect that our approach and findings would also apply to defenses other than adversarial training.

\subsection{Fairness and Adversarial Robustness}

Algorithmic fairness, in general, and fairness in \ml{}, specifically, have been extensively studied in recent years~\cite{verma2018fairness}.
Researchers have defined different fairness notions between individuals and groups, ranging from precision parity to recall parity~\cite{verma2018fairness},
studied when notions can or cannot be satisfied simultaneously (e.g.,~\cite{chouldechova2017fair}), and proposed means to promote different notions of fairness (e.g.,~\cite{corbett2017algorithmic, singh2024representation}).
In addition to ethical considerations that often motivate fairness, another key motivation behind fairness is improving the general performance of the system (e.g., benign accuracy), which sometimes stems as a side effect of improving fairness~\cite{karkkainenfairface}.

Prior work has mainly focused on studying fairness in the context of benign inputs (e.g.,~\cite{agarwal2018reductions, hashimoto2018fairness, karkkainenfairface}). 
However, a line of work has studied fairness in the context of adversarial robustness---particularly, disparities in adversarial robustness between different (source) classes and groups~\cite{nanda2021fairness, xu2021robust}.
While defenses against adversarial examples such as adversarial training significantly improve robust accuracy, they do not guarantee fairness across classes or demographic groups. %
In high-stakes applications such as face recognition, this limitation is particularly concerning: a system may appear robust overall, yet still perform disproportionately worse on certain classes or subpopulations, exposing vulnerable groups to higher risks of misidentification or security breaches.  
In concurrent work, Nanda et al.~\cite{nanda2021fairness} and Xu et al.~\cite{xu2021robust} noticed that certain classes or groups may have significantly lower robust accuracy than others.
These efforts highlighted the shortcomings of evaluating fairness only under benign conditions, and reinforced the need for defenses that explicitly address both robustness and fairness under attack.  
As noted by Rosenberg et al.~\cite{rosenberg2023fairness}, similar disparities are also exhibited by face-recognition models where face-obfuscation with $\ell_p$-norm bounded perturbations to attain privacy are less effective for minority groups.

\paragraph{Improving Fair Adversarial Robustness.}
Several methods have emerged to tackle fairness in adversarial settings~\cite{jin2025enhancing, li2023wat, ma2022fat, sun2023improving, wei2023cfa, xu2021robust, zhang2024towards}.
In particular, 
\emph{Fair Robust Learning (FRL)}~\cite{xu2021robust} aims to improve worst-class robust accuracy through a min–max framework that reweights adversarial samples and adjusts attack strengths during training.
\emph{Balance Adversarial Training (BAT)}~\cite{sun2023improving} tackles fairness by encouraging uniformity in predicted classes and equalizing attack difficulty across source classes.
\emph{Fairness-Aware Adversarial Learning (\faal)}~\cite{zhang2024towards} improves worst-class robustness by dynamically reweighting samples and using a form of so-called Distributionally Robust Optimization (DRO) to optimize worst-case performance. 
Last, \emph{Confusional Spectral Regularization (\specnorm)}~\cite{jin2025enhancing}
penalizes the spectral norm of the confusion matrix to bound the worst-case class-level robust error. %
Similar to \specnorm{}, \defense{} optimizes particular properties of the confusion matrix on adversarial examples (\secref{sec:technical_approach}).
However, %
\defense{} departs from norm-based constraints and instead focuses on directional imbalances in misclassification rates.
\specnorm{} and \faal{} are considered the leading approaches for enhancing fair source-class adversarial robustness, thus we include them in our experiments (\secref{sec:experiment_setup}).

In contrast to prior work, \defense{} aims to improve fair adversarial robustness through novel means, by optimizing symmetry---while this property sometime naturally arises, misclassification patterns of adversarial examples between most class pairs are often asymmetric~\cite{mei2023towards}.
Moreover, in this work we identify disparities in target-class adversarial robustness as a key limitation of existing techniques and show that \defense{} helps make it less severe. 
Last, previous work has mainly focused on predefined groups (e.g., genders or races), while we show how to improve fair adversarial robustness for any arbitrary group---addressing an issue that is known to be computationally challenging for certain notions of fairness.

We also note that previous defenses aiming to promote fairness and robustness have mainly conducted evaluations on object-classification benchmarks (e.g., CIFAR-10 and CIFAR-100) where robustness and fairness have little-to-no security or ethical implications.
Moreover, they considered unrealistic adversaries that may fail to harm integrity under real-world constraints.
In contrast, in this work, we mainly focus on high-stakes settings by adapting prior techniques and evaluating them along with \defense{} on face-recognition models against realistic eyeglass attacks.

\section{Threat Model}
\label{sec:threat-model}
\label{sec:threat}

We consider settings where adversaries produce adversarial examples to mislead \ml{}-based image classifiers, particularly, ones based on neural networks.
The adversaries we consider have white-box access to models, enabling them to efficiently produce worst-case perturbations using their knowledge of model weights, in contrast to black-box counterparts, which often slower and achieve lower attack success~\cite{Croce20Auto}.
Moreover, we consider adversaries launching either untargeted or targeted attacks.
Furthermore, we focus on adversaries that use realistic attacks against face-recognition (specifically, using the eyeglass attack~\cite{sharif2016accessorize}).
However, for better compatibility with prior work on fair adversarial robustness~\cite{jin2025enhancing, zhang2024towards}, we also study adversaries that produce imperceptible adversarial perturbations bounded in $\ell_p$-norm against object recognition. 
In both cases, the adversary's goal is to maximize their success rate---i.e., achieving any misclassification in the case of untargeted attacks, or obtaining misclassification to target classes in the case of targeted attacks---thus maximally harming the integrity of the \ml{}-based system.

Contrastively, the defender aims to prevent 
attempts to mislead the \ml{} model to the extent possible (i.e., \emph{maximizing robust accuracy}) while \emph{maximizing benign accuracy} on clean inputs.
Specifically, the defender induces adversarial robustness at training time, through a form of adversarial training.
Crucially, in this work, we place strong emphasis on fair adversarial robustness.
Specifically, additionally to deterring attacks and correctly classifying benign samples, our defender seeks to \emph{maximize fair source and target adversarial robustness, both at the individual class and group levels}.

\section{Sy-FAR's Technical Approach}
\label{sec:technical_approach}

Our proposed symmetry-based regularization builds upon confusion-matrix analysis, which has been recognized as a foundation for fairness evaluation~\cite{gursoy2022equal}.
Here, however, we focus on confusion matrices over adversarial examples.
We follow a standard adversarial training framework, where the objective balances benign accuracy and adversarial robustness, and extend it with a fairness regularizer. 
Unlike earlier fairness-oriented regularizers that rely on specialized optimization schemes ~\cite{jin2025enhancing, sun2023improving, xu2021robust}, \defense{} is conceptually straightforward yet effective: it introduces a differentiable regularization term that explicitly enforces symmetry between class pairs, thereby counteracting asymmetry in adversarial settings. 
The intuition is that if examples of class \(i\) can be perturbed to resemble class \(j\), then the reverse should hold as well---class similarity is fundamentally a symmetric relation. 
Framing fairness in this way provides a natural and tractable objective even when considering unbalanced or biased datasets.
Moreover, the approach is computationally lightweight, as evaluating the regularizer requires only a single pass over the confusion matrix, making it well-suited for scalable adversarial training.

Next, we provide intuition for why symmetry helps advance fair adversarial robustness (\secref{sec:syfar:intuition}), describe the symmetry regularizer (\secref{sec:tech:symreg}), and present \defense{}'s overall objective and algorithm (\secref{sec:objectives}).

\subsection{Intuition: Why Symmetry?}
\label{sec:syfar:intuition}

\begin{figure}[t!]
    \centering
    \begin{subfigure}[t]{0.20\textwidth}
        \centering
        \includegraphics[width=0.9\linewidth]{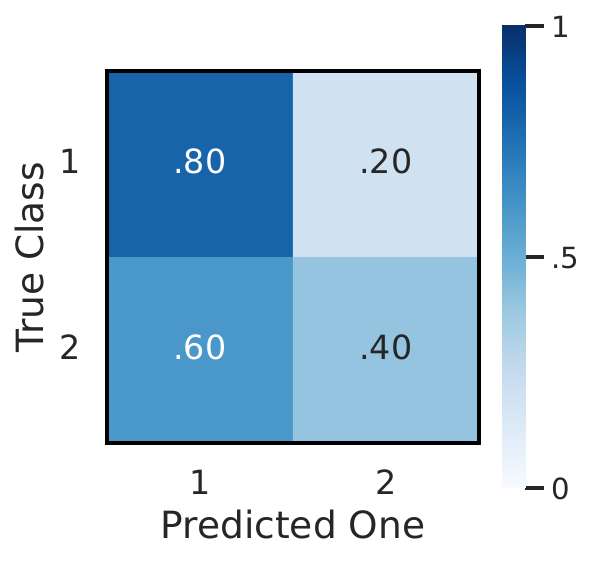}
        \caption{Asymmetric confusion}
        \label{fig:asymmetric_conf}
    \end{subfigure}
    \hfill
    \begin{subfigure}[t]{0.20\textwidth}
        \centering
        \includegraphics[width=0.9\linewidth]{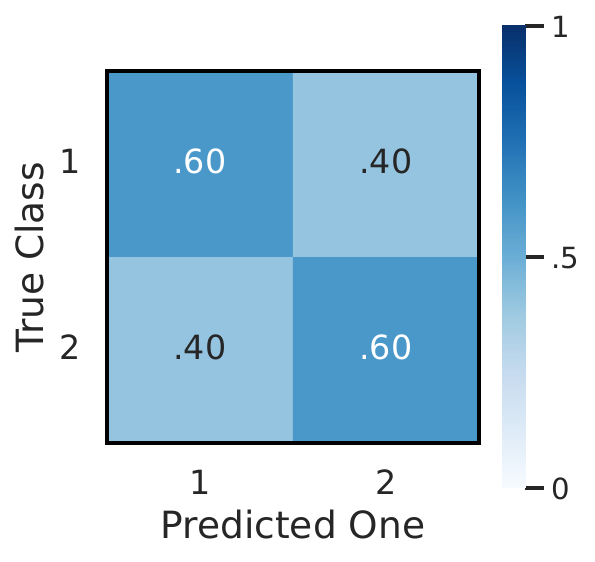}
        \caption{Symmetric confusion}
        \label{fig:symmetric_conf}
    \end{subfigure}
    \hfill
    \caption{An illustration of asymmetric vs.\ symmetric confusion matrices.}
    \label{fig:confusion_comparison}
\end{figure}

\figref{fig:confusion_comparison} illustrates two confusion matrices over adversarial examples.
While one matrix is asymmetric and another is symmetric, both represent models with an overall 60\% robust accuracy.
Note, however, that the model with the asymmetric confusion matrix exhibits unfair source- and target-class adversarial robustness: class 2 has robust accuracy lower by 40\% than class 1, and adversarial examples have 40\% higher likelihood of being classified into class 1 than into class 2.
In contrast, the model with the symmetric confusion matrix achieves fair source- and target-class adversarial robustness (both classes have identical robust accuracy and they are equally likely of being selected as output of successful adversarial examples).
Thus, overall, this illustrative example shows that if we are able to maintain (or increase) robust accuracy while improving symmetry, it is possible to promote fair adversarial robustness.
In our experiments (\secsref{sec:experiment_results}{sec:fairness_subgrp}), we find that symmetry regularization indeed preserves or boosts robust accuracy, and, as expected, also improved fair adversarial robustness from source- and target-class perspectives.

\subsection{Symmetry Regularization}
\label{sec:tech:symreg}

Confusion matrices provide a natural way to visualize how predictions are distributed across classes. In a standard (``hard'') confusion matrix, each entry \(C_{ij}\) counts how many examples from class \(i\) are predicted as class \(j\). 
The diagonal entries represent correct classifications, while the off-diagonal entries correspond to misclassifications. 
A symmetric confusion matrix, \(C\), is one where \(C_{ij} = C_{ji}\), meaning that if samples of class \(i\) are often confused with class \(j\), the reverse confusion occurs at a similar rate. 
By contrast, asymmetry arises when one direction dominates, e.g., class \(i\) is frequently misclassified as \(j\) but not vice versa. 
Such directional imbalances reflect fairness issues: one class disproportionately ``loses'' to another, making it more vulnerable under adversarial perturbations (as illustrated in \secref{sec:syfar:intuition}).

Producing standard confusion matrices requires computing predicted class labels. 
However, since we cannot use hard predictions in gradient-based optimization due to non-differentiability, analogous to how the benign loss relies on softmax probabilities in lieu of discrete labels, we adopt a similar approach here. 
Specifically, we define a \emph{soft confusion matrix}, where each entry \(C_{ij}\) represents the expected probability of samples from class \(i\) being predicted as class \(j\). 
This construction leverages the softmax outputs of the model, making it directly compatible with gradient-based optimization.

\paragraph{Soft Confusion Matrices From Probabilities.}
Let $K$ be the number of classes. For a minibatch of $B$ adversarial examples $\{(x_b, y_b)\}_{b=1}^B$, where $x_b$ denotes the input image and $y_b$ its ground-truth label, with model logits $z_b \in \mathbb{R}^K$, we first compute the softmax probabilities $p_b = \mathrm{softmax}(z_b)$ for each sample. To estimate the confusion matrix in a differentiable way, we form $\widehat{C} \in \mathbb{R}^{K \times K}$ by accumulating these probabilities according to ground-truth labels: for each class $i$, we sum (or equivalently average) the probability vectors $p_b$ over all samples with $y_b = i$, and assign the result to row $i$ of $\widehat{C}$. In this construction, each row corresponds to a true class, while each column reflects the (soft) predicted distribution for that class.
We also track the number of samples per class using a count vector $n \in \mathbb{R}^K$. 
Starting with $\widehat{C}$ initialized with zeros, after accumulating the soft predictions, we normalize each row by the corresponding count to obtain a final soft confusion matrix $C \in \mathbb{R}^{K \times K}$. 
In other words, for each prediction $p_b$, we accumulate $\widehat{C}_{y_b,:} \leftarrow \widehat{C}_{y_b,:} + p_b^\top$. 
Subsequently, we normalize each row $i$ by $n_i$, the number of samples in class $i$ (i.e., we set $C_{i,:}\leftarrow\frac{\widehat{C}_{i,:}}{\max(n_i, 1)}$).
Here, $\max(n_i,1)$ prevents division by zero in case a class is missing from the minibatch.
Eventually, each cell $C_{ij}$ estimates the conditional class distribution $\mathbb{P}(\widehat{y}=j \mid y=i)$ using soft predictions, where $\widehat{y}$ denotes the predicted class of the adversarial example.

\paragraph{Pairwise Asymmetry Penalty.}
\defense{}'s goal is to reduce directional bias between pairs of classes: if samples from class $i$ are often misclassified as class $j$, then misclassifications in the reverse direction should occur at a similar rate. 
To capture this, we define a penalty for each class pair, as follows.  
For a class pair $(i,j)$ with $i<j$, denote $a = C_{ij}$ and $b = C_{ji}$. 
We then penalize \emph{relative} asymmetry, scaled by the \emph{total misclassification mass}:
\begin{equation}
\label{eq:pair-penalty}
\Penalty_{i,j}
\;=\;
\frac{|a-b|}{a+b+\epsilon}\cdot(a+b),
\end{equation}
where $\epsilon$ is a small constant for numerical stability, set in proportion to the symmetry-loss weight hyperparameter (default $\epsilon = \tfrac{1}{K}$).  
The first factor $\tfrac{|a-b|}{a+b+\epsilon}$ measures the degree of directional imbalance (zero when $a \approx b$), while $(a+b)$ scales the penalty more strongly for class pairs that the model frequently confuses (i.e., large off-diagonal entries). This scaling is important: without it, even negligible cases—such as one misclassification in one direction and none in the other—would contribute a large relative imbalance despite being practically insignificant. By weighting with $(a+b)$, the loss emphasizes meaningful asymmetries from frequent confusions while downplaying spurious imbalances from rare or single-sample errors.

The total symmetry loss then aggregates penalties across all unordered class pairs: 
\begin{equation}
\label{eq:sym-loss}
\SymLoss(C)
\;=\;
\sum_{1 \le i < j \le K} \Penalty_{i,j}.
\end{equation}

\paragraph{Interpretation and Differentiability.}
The symmetry regularizer is minimal when $C_{ij}=C_{ji}$ for all $i\neq j$, and it increases smoothly with both (i) the magnitude of asymmetry and (ii) the amount of mutual confusion. 
This design directly targets directional bias (e.g., $A\!\to\!B \gg B\!\to\!A$) while de-emphasizing rare, noisy interactions. 
Moreover, all steps from logits $\to$ softmax $\to$ row-normalized $C$ $\to$ \eqnref{eq:sym-loss} involve mostly continuous and differentiable operations (addition, multiplication, absolute value, %
etc.), making the loss fully differentiable.

\subsection{Training With Symmetry Regularization}
\label{sec:objectives}
\label{sec:tech:alg}

\paragraph{Loss Composition.}
We follow the adversarial training framework, where the objective balances benign accuracy and adversarial robustness, and extend it with our symmetry regularizer. Specifically, the total loss combines clean loss, adversarial loss, and symmetry loss:
\begin{equation*}
\label{eq:total-loss}
\mathcal{L}
\;=\;
\lambda_{\mathrm{clean}}\ \mathcal{L}_{\mathrm{CE}}(x,y)
\;+\;
\lambda_{\mathrm{adv}}\ \mathcal{L}_{\mathrm{CE}}(x^{\mathrm{adv}},y)
\;+\;
\lambda_{\mathrm{sym}}\ \SymLoss(C),
\end{equation*}
where $\mathcal{L}_{\mathrm{CE}}$ is the cross-entropy loss on clean inputs $(x,y)$ and adversarial inputs $(x^{\mathrm{adv}},y)$; $C$ is the soft confusion matrix of the adversarial batch; and $\lambda_{\mathrm{clean}},\lambda_{\mathrm{adv}},\lambda_{\mathrm{sym}}>0$ are scalar hyperparameters balancing the three losses.

\algoref{alg:sym-train} summarizes the training procedure, where each minibatch involves a clean forward pass, adversarial example generation, construction of the soft confusion matrix, and the computation of the symmetry loss. 
The resulting objective integrates these components and updates the model parameters accordingly via gradient descent.

\begin{algorithm}[t!]
\caption{\defense{}'s Algorithm}
\label{alg:sym-train}
\begin{algorithmic}[1]
\REQUIRE dataset $\mathcal{D}$, model $f_\theta$, attack $\mathcal{A}$, weights $(\lambda_{\mathrm{clean}},\lambda_{\mathrm{adv}},\lambda_{\mathrm{sym}})$
\FOR{each minibatch $(x,y)=\{(x_b,y_b)\}_{b=1}^B$}
  \STATE \textbf{Clean pass:} $z \!\leftarrow\! f_\theta(x)$,\quad $\mathcal{L}_{\mathrm{clean}}\!\leftarrow\!\mathcal{L}_{\mathrm{CE}}(z,y)$
  \STATE \textbf{Adversarial examples:} $x^{\mathrm{adv}}\!\leftarrow\!\mathcal{A}(f_\theta,x,y)$
  \STATE \textbf{Adversarial pass:}
  \STATE ~\ ~\ $z^{\mathrm{adv}}\!\leftarrow\! f_\theta(x^{\mathrm{adv}})$,\quad $p\!\leftarrow\!\mathrm{softmax}(z^{\mathrm{adv}})$
  \STATE \textbf{Soft confusion:} build $C$ per \secref{sec:tech:symreg}
  \STATE \textbf{Symmetry loss:} $\SymLoss(C)\!\leftarrow\!\sum_{i<j}\frac{|C_{ij}-C_{ji}|}{C_{ij}+C_{ji}+\epsilon}(C_{ij}+C_{ji})$
  \STATE \textbf{Adversarial loss:} $\mathcal{L}_{\mathrm{adv}}\!\leftarrow\!\mathcal{L}_{\mathrm{CE}}(z^{\mathrm{adv}},y)$
  \STATE \textbf{Total loss:} $\mathcal{L}\!\leftarrow\!\lambda_{\mathrm{clean}}\mathcal{L}_{\mathrm{clean}}+\lambda_{\mathrm{adv}}\mathcal{L}_{\mathrm{adv}}+\lambda_{\mathrm{sym}}\SymLoss(C)$
  \STATE \textbf{Update:} backpropagate $\nabla_\theta \mathcal{L}$ and step optimizer
\ENDFOR
\end{algorithmic}
\end{algorithm}

\paragraph{Complexity.}
Computing $\SymLoss(C)$ involves comparing each unordered class pair $(i,j)$ exactly once, resulting in \textbf{$O(K^2)$} operations per batch for $K$ classes. 
The computation consists only of simple element-wise differences and sums over the confusion matrix, which are fully vectorizable on a GPU. 
In typical regimes, the added cost is negligible compared to the standard forward and backward passes, making \defense highly efficient even for tasks with many classes.

\section{Experimental Setup}
\label{sec:experiment_setup}

This section outlines the baseline methods, datasets, models, attack settings, and evaluation metrics used throughout our experiments.
A high-level summary of the experimental configurations is provided in \tabref{tab:narrow-exp-settings}. %
We present the results in \secref{sec:experiment_results}, where we highlight consistent gains in robustness and fairness over all baselines.

\paragraph{Datasets and Models.}
To evaluate fair adversarial robustness under realistic settings, we conduct experiments on three variations of the \pubfig face-recognition dataset~\cite{attribute_classifiers}. 
First, we use a gender-balanced subset of {\pubfig} featuring ten celebrity identities (five female, five male), with approximately 300 images per identity.\footnote{Males: Antonio Banderas; Colin Powell; Hugh Grant; John Travolta; Will Smith.  Females: Jennifer Lopez; Meryl Streep; Oprah Winfrey; Reese Witherspoon; Tyra Banks.}
This selection was motivated by the need to control for demographic imbalance: by ensuring equal male–female representation, differences in misclassification rates can be attributed to model behavior rather than dataset bias.
As our main model, we use the {\vgg} architecture~\cite{simonyan2015very}, a well-established convolutional neural network (CNN).
Second, we modify the dataset to {\pubfigsiblings} by replacing two female identities in \pubfig with a pair of visually similar female siblings.\footnote{We replace Meryl Streep and Oprah Winfrey with Dakota and Elle Fanning, while retaining the remaining identities from \pubfig.}
Doing so preserves the gender balance (five male, five female identities) and the dataset size ($\sim$300 images per identity, as in \pubfig), with the two replaced identities selected at random to maintain diversity. 
\pubfigsiblings{} setup stress-tests the system's ability to ensure fairness not only across gender but also among individuals with high visual similarity---a critical scenario for symmetry-aware fairness evaluation.
Finally, we test generalization beyond CNNs by applying \defense to a {Vision Transformer (\vit)} trained on {\pubfigvit}.

We pre-process all face images using the PyTorch-based FaceX-Zoo toolkit~\cite{wang2021facex}, which provides standard pipelines for face alignment and cropping. 
We adopt an 80/10/10 split for training, validation, and testing, a standard ratio that provides sufficient training diversity while keeping held-out data for reliable evaluation. 
Across all experiments, we perform ten independent repetitions per setup and report averaged results, where each training method is evaluated under its optimized hyperparameters \ifArxiv(see \appref{app:training} for details on the training method).
\fi

All methods are fine-tuned on top of adversarially trained models, where we empirically find that five epochs suffice for the \pubfigsiblings and \pubfigvit{} settings, while ten epochs yielded optimal results for \pubfig.

\begin{table}[t!]
\centering
\scriptsize
\begin{tabularx}{\columnwidth}{@{}l l l l X@{}}
\toprule
\textbf{Data} & \textbf{Model} & \textbf{Def.} & \textbf{Eval} & \textbf{Train} \\
\midrule
\pubfig & \vgg & ROA & \makecell[l]{Eyeglass \\ U+T} & 10e FT \\
\makecell[l]{\pubfig+ \\ Sibs} & \vgg & ROA & \makecell[l]{Eyeglass \\ U+T} & 5e FT \\
\pubfig & \vit & ROA & \makecell[l]{Eyeglass \\ U+T} & 5e FT \\
CIFAR-10 & PreAct-ResNet18 & \makecell[l]{PGD-$\ell_\infty$ \\ ($\epsilon=8/255$, \\ $\alpha=2/255$, 10 iters)} & AutoAttack & 200e Scratch \\
CIFAR-100 & PreAct-ResNet18 & \makecell[l]{PGD-$\ell_\infty$ \\ ($\epsilon=8/255$, \\ $\alpha=2/255$, 10 iters)} & AutoAttack & 200e Scratch \\
\bottomrule
\end{tabularx}
\caption{Summary of experimental setups. U and T stand for untargeted and targeted attacks, respectively. FT stands for fine-tuning epochs, and Scratch denotes training from scratch.}
\label{tab:narrow-exp-settings}
\end{table}

To enable direct comparison with prior work that focused primarily on CIFAR-10 and CIFAR-100~\cite{krizhevsky2009learning}, and to demonstrate that our results generalize beyond face recognition, we also evaluate \defense on these benchmarks ~\cite{krizhevsky2009learning} with Preact-ResNet18 model, as commonly used in the adversarial robustness and fairness literature~\cite{xu2021robust, sun2023improving, zhang2024towards, jin2025enhancing}. 
Since these datasets do not capture the real-world fairness challenges central to our work (e.g., demographic balance or sibling similarity), we defer the detailed results in \appref{app:cifar-results}.
For both CIFAR-10 and CIFAR-100 with PreAct-ResNet18, all models are adversarially trained from scratch for {200 epochs}.

\paragraph{Defenses and Baselines.}  
All face recognition experiments adopt the Rectangular Occlusion Attack (ROA) defense, following the FACESEC framework~\cite{tong2021facesec}.  
ROA simulates physical-world patch-based occlusions by performing an exhaustive search over fixed-size rectangular regions to find the most damaging location, then runs constrained attack 
within that region.  
We use FACESEC's default parameters: rectangle size 70${\times}$70, stride (10,10), step size of $\alpha$=20, and 100 attack %
iterations, with occlusion color initialized to mid-gray ($255/2$ in pixel space).  
These settings give the best results in our experiments, offering strong robustness while maintaining competitive clean accuracy, and we adopt them consistently across all face-recognition setups. 

On CIFAR-10 and CIFAR-100, we adversarially train the PreAct-ResNet18 model using PGD-$\ell_\infty$~\cite{madry2018towards}, 
with perturbation budget $\epsilon$=8/255, step size $\alpha$= 2/255, and ten iterations per step.  
This configuration is common in recent fair adversarial robustness studies, including \faal{}~\cite{zhang2024towards} and \specnorm{}~\cite{jin2025enhancing}, and thus provides a standard benchmark for comparison.

\paragraph{Fine-tuning for Enhancing Robust Fairness.} 
Since our focus is specifically on the challenge of fair adversarial robustness, we build on adversarially trained models that already achieve a reasonable level of average robustness.
This strategy is consistent with prior work, such as \faal{}~\cite{zhang2024towards}, which fine-tunes on top of adversarial training to further improve fairness-oriented objectives. 
In this context, we investigate whether fairness can be effectively improved by fine-tuning robust models, rather than retraining entirely from scratch.
Our experiments indicate that adversarial fine-tuning is a more efficient approach: it enhances both fairness and robustness while avoiding the overhead of training from scratch.

Accordingly, we adopt fine-tuning as our primary setting for face-recognition tasks. 
For completeness, in the object recognition benchmarks (CIFAR-10/100), we explored both training from scratch and fine-tuning, and found their performance to be nearly identical; reported numbers (\appref{app:cifar-results}) correspond to the best-performing configuration.

\begin{figure}[t!]
    \centering
    \includegraphics[width=\columnwidth]{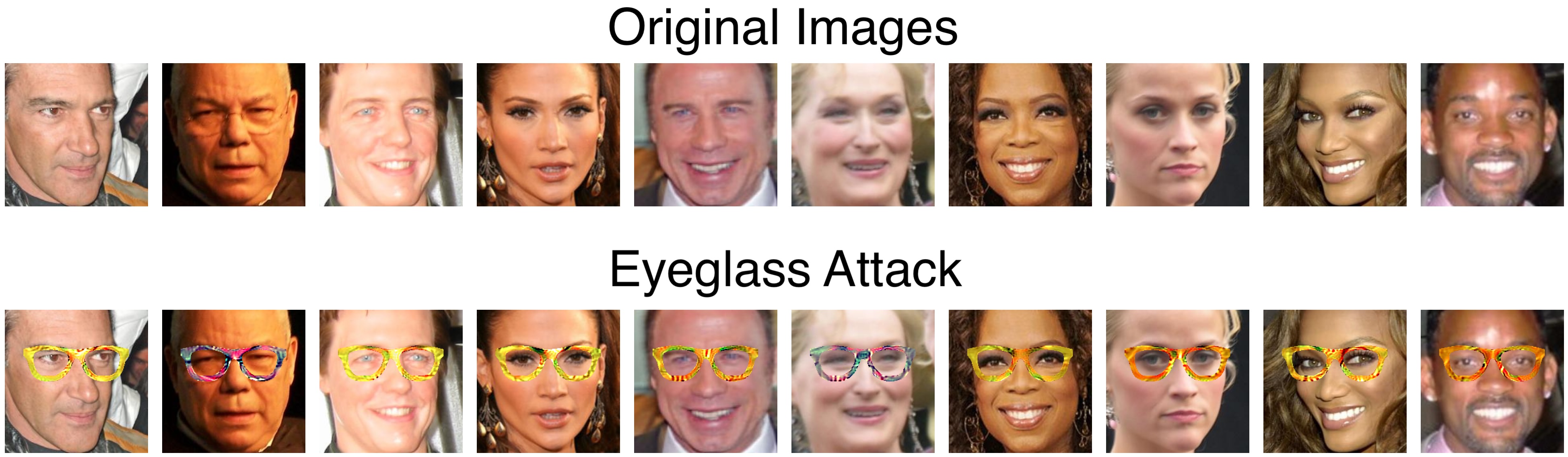}
    \caption{
    Illustration of the eyeglass attack on PubFig images~\cite{attribute_classifiers}. 
    Original face images are at the top, and images perturbed with adversarial eyeglasses are at the bottom~\cite{sharif2016accessorize}. 
    The attack only modifies pixels within the eyeglass region, producing perturbations that mislead face recognition. 
    }
    \label{fig:pubfig_glass}
\end{figure}

\paragraph{Attacks.}  
We evaluate the face-recognition models under the realistic eyeglass attack~\cite{sharif2016accessorize}. 
The attack overlays a fixed eyeglass-frame mask onto the face image and perturbs pixel values only within the masked region. %
The perturbation is initialized to a fixed color maximizing the (mis)classification loss.
Adversarial examples are then refined through iterative gradient-based updates constrained to the eyeglass region, with momentum to stabilize optimization.  
We test a wide range of iterations capturing both weak and strong attack regimes;
results are reported for the strongest attack configuration (with 300 iterations).
Moreover, we use both the untargeted (primarily) and targeted variants of the attack.
This setting reflects realistic threats where adversarial glasses can be realized by adversaries to mislead face recognition, making it a natural testbed for fairness evaluation.
\figref{fig:pubfig_glass} depicts attack examples with images from \pubfig{}, 
demonstrating how visually plausible perturbations can drastically impact recognition outcomes.

On CIFAR-10 and CIFAR-100, evaluation is performed using AutoAttack~\cite{Croce20Auto}, a parameter-free ensemble of four strong adversarial attacks: (1) APGD-CE, (2) APGD-DLR~\cite{Croce20Auto}, (3) FAB~\cite{croce2020minimally}, and (4) Square Attack~\cite{andriushchenko2020square}.  
Consistent with prior fairness-aware adversarial training approaches, including \faal{}\cite{zhang2024towards} and \specnorm{}\cite{jin2025enhancing}, we adopt this evaluation setup as it provides a reliable benchmark for comparison.

\paragraph{Evaluation Metrics.}  
We assess accuracy, robustness, and fairness with multiple metrics: 
\begin{itemize}[leftmargin=10pt, itemsep=0pt]
    \item \textbf{Benign Accuracy:} The standard classification accuracy under clean (unperturbed) inputs. This serves as a sanity check for overall model performance and ensures that robustness or fairness improvements do not come at the expense of degraded clean accuracy. 

    \item \textbf{Robust Accuracy:} The average accuracy under adversarial attack (untargeted or targeted, depending on the setup). This is the primary measure of adversarial robustness and is widely used in the literature (e.g.,~\cite{madry2018towards, engstrom2018evaluating, kannan2018adversarial, zhang2020mixup, lee2020adversarial}).

    \item \textbf{Source-class Fairness:} We also evaluate fair adversarial robustness with respect to the source class. %
    We adopt two standard metrics from the literature to quantify whether robustness is distributed evenly across classes~\cite{xu2021robust,sun2023improving,zhang2024towards,jin2025enhancing}:
    \emph{(1) Min (Worst-Class) Accuracy} reports the lowest diagonal entry of the row-normalized confusion matrix, capturing the most vulnerable class under attack; and
    \emph{(2) Accuracy Gap} measures the difference between the maximum and minimum values on the confusion matrix's diagonal, measuring disparity across classes.  

    \item \textbf{Symmetry-based Metrics:} These metrics quantify directional bias in misclassifications using the hard confusion matrix ($C$): %
    \emph{(1) Max Asymmetry Gap} measures the largest imbalance between two classes, $\max_{i,j}|C_{ij} - C_{ji}|$; and
    \emph{(2) Symmetry Loss} is our regularization loss, $\mathcal{L}_{\mathrm{sym}}(C)$, from Eq.~\eqref{eq:sym-loss}.  
    These metrics reveal whether class confusions are bidirectionally balanced (i.e., misclassification of adversarial examples from $i\!\to\!j$ occurs as often as from $j\!\to\!i$). Lower values indicate more symmetric and fair behavior.
    \label{sec:metrics}

    \item \textbf{Target-class Fairness:} 
    Here, metrics evaluate fairness from the target-class perspective.  
    For each target class $j$, we compute the fraction of total misclassifications that are directed into $j$:  
    \[
    T_j = \frac{\sum_{i \neq j} C_{ij}}{\sum_{i \neq j}\sum_{k \neq i} C_{ik}}.
    \]  
    This \emph{target confusion share} indicates how often class $j$ becomes the destination of misclassifications.
    Based on the fraction of total misclassifications, we define three quantities:
    \emph{(1)} $\mathrm{TgtMax} = \max_j T_j$: here, lower is better, since this quantity reveals whether one class receives a disproportionately large share of errors (bias); and
    \emph{(2)} $\mathrm{TgtMin} = \min_j T_j$: here, higher is better, since very low values imply that some classes are almost never predicted, rendering them more advantaged than others.
    \emph{(3)} $\mathrm{TgtStd} = \mathrm{Std}_j(T_j)$: the standard deviation across all $\{T_j\}$ values, capturing how evenly misclassifications are distributed among target classes (lower indicates greater fairness).
\end{itemize}

\section{Class-Level Fairness: Experiment Results}
\label{sec:experiment_results}

This section presents our experimental results, demonstrating that, by improving symmetry between individual classes (\secref{sec:symmetry_loss}), \defense{} improves robust fairness from source- and target-class perspectives while mostly preserving or improving benign and robust accuracy compared to baseline approaches (\secsref{sec:source_class}{sec:target_class}). 
Our experiments also demonstrate faster run time compared to baselines (\secref{sec:runtime}), and higher stability across runs (\secref{sec:stability}).
Overall, these results provide a comprehensive view of how symmetry-based regularization advances robustness and fairness simultaneously across diverse settings (datasets, model architectures, etc.).

In this section, we focus on untargeted eyeglass attacks against the face-recognition models.
We note that the results on CIFAR-10 and CIFAR-100, reported in \appref{app:cifar-results} follow consistent trends with those obtained on the face-recognition models. 
Similarly, results for the \emph{targeted} eyeglass attack (\appref{app:tgtd_attack})
and powerful \emph{face-mask attacks}~\cite{tong2021facesec} (\appref{sec:strong_attacks}) also corroborate our main findings. 

\subsection{Symmetry Loss}
\label{sec:symmetry_loss}
We begin by evaluating whether explicitly optimizing for symmetry in parallel to other objectives (as described in~\secref{sec:experiment_setup}) indeed improves symmetry on the adversarial examples produced from the test samples of the dataset.
\tabref{tab:sym_untgtd} summarizes the results for the face-recognition datasets and models, using the untargeted eyeglass attack to produce adversarial examples. 
It can be seen that, compared to vanilla adversarial training, \faal{} and \specnorm{}, both yield better symmetry (i.e., lower asymmetry gap and symmetry loss), showcasing the connection between fairness and symmetry---optimizing for fair source-class robustness directly improves symmetry even without explicitly optimizing for symmetry.
In comparison, \defense{} has the largest impact on symmetry, achieving the lowest asymmetry gap and symmetry loss across all methods evaluated.
\defense{}'s gains are particularly pronounced in the \pubfigsiblings setting, where visually similar identities exacerbate directional misclassifications. 

\begin{figure*}[t!]
  \centering
  \begin{subfigure}{0.69\columnwidth}
    \centering
    \includegraphics[width=\linewidth]{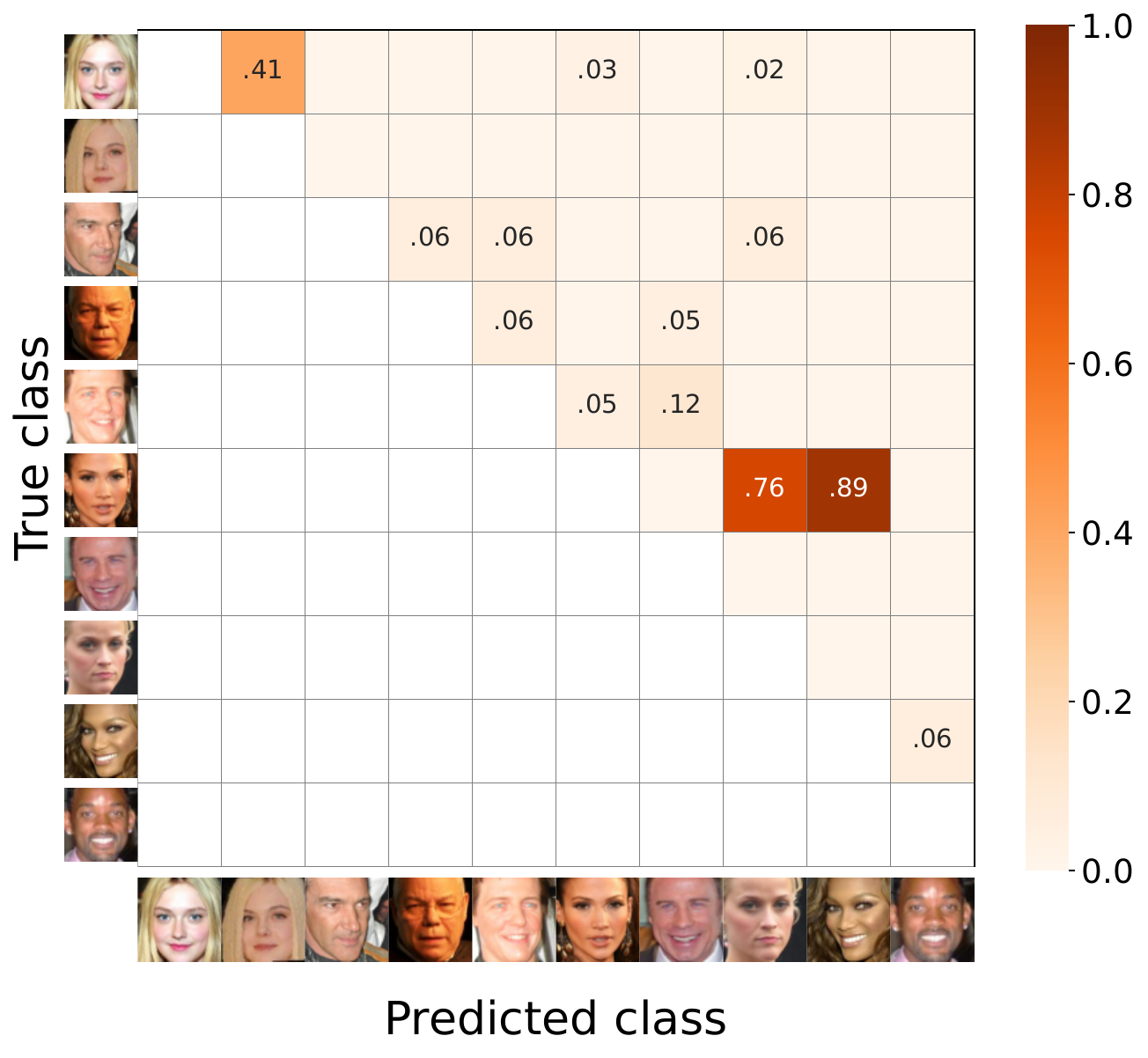}
    \caption{\faal}
  \end{subfigure}
  \hfill
  \begin{subfigure}{0.69\columnwidth}
    \centering
    \includegraphics[width=\linewidth]{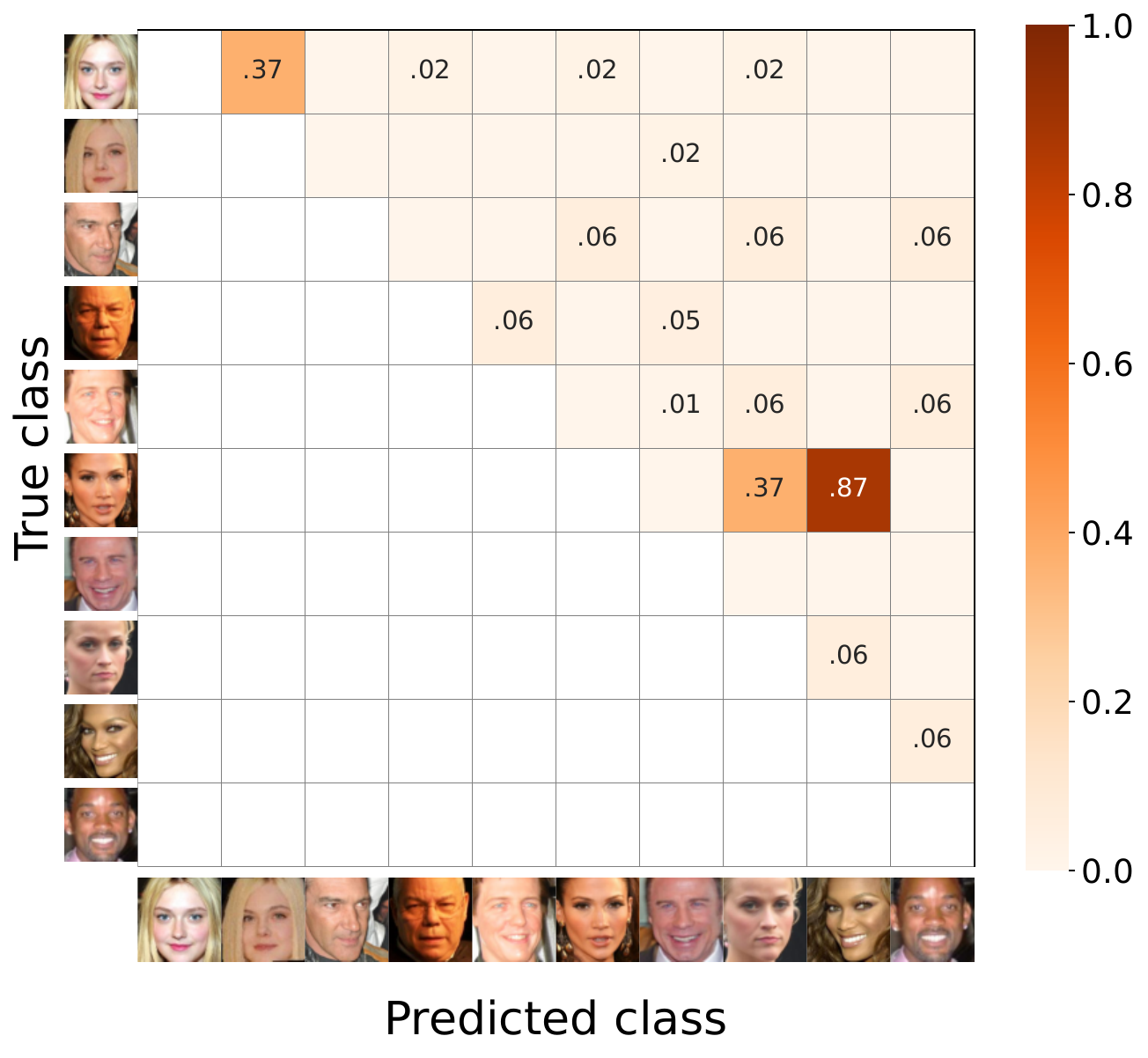}
    \caption{\specnorm}
  \end{subfigure}
  \hfill
  \begin{subfigure}{0.69\columnwidth}
    \centering
    \includegraphics[width=\linewidth]{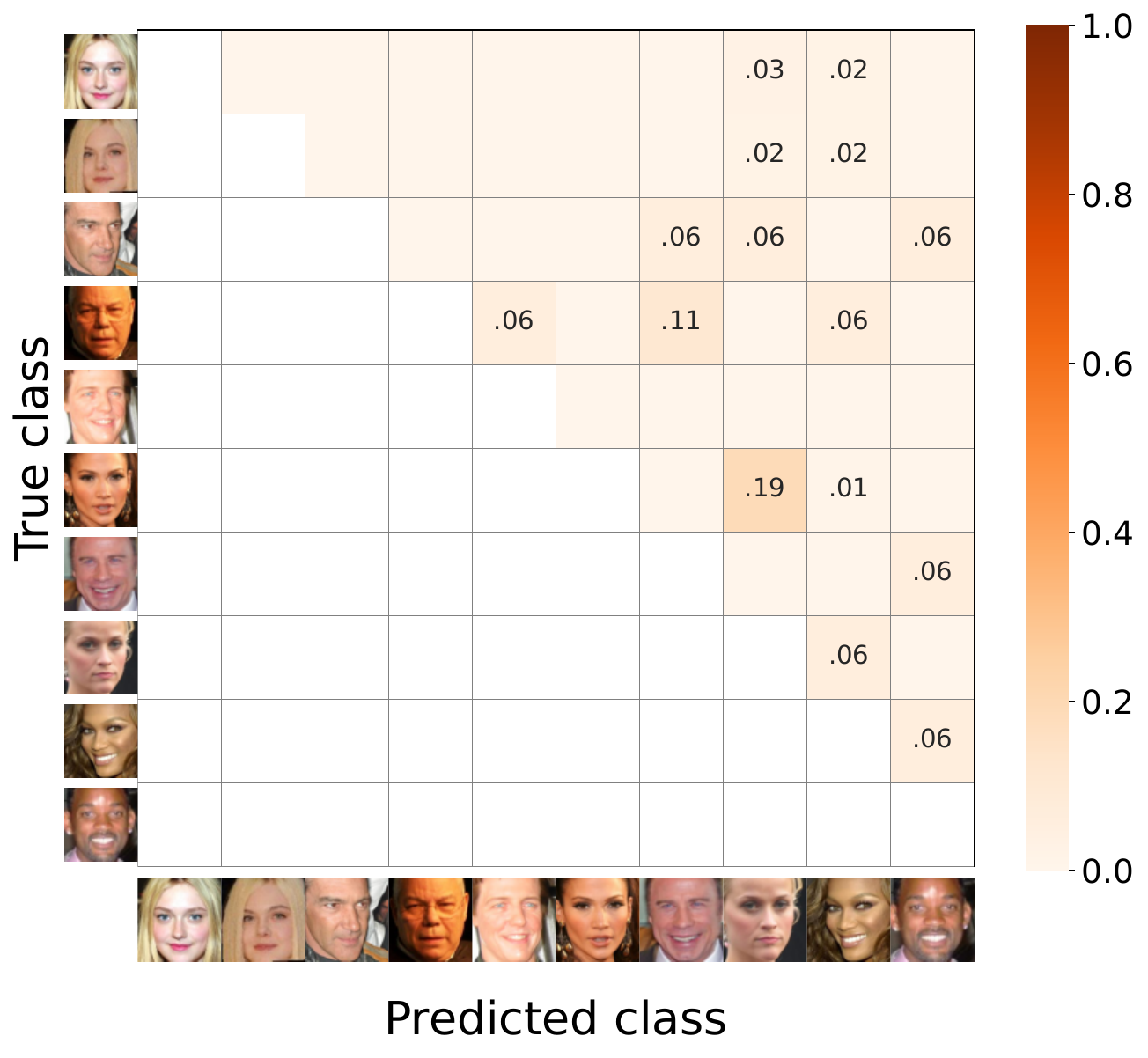}
    \caption{\defense{}}
  \end{subfigure}
    \caption{
    Asymmetry heatmaps on \pubfigsiblings, using the untargeted eyeglass attack to create adversarial examples. 
    Each cell $(i,j)$ in the upper triangle reports the \emph{Asymmetry Gap}, i.e., $|C_{ij}-C_{ji}|$ (see \secref{sec:metrics}). 
    Darker regions indicate stronger directional bias, where adversarial examples from one class are more likely to be classified into the other class than vice versa. We show representative heatmaps from individual randomly selected runs out of the ten repetitions.
}
    \label{fig:symmetry-all}
\end{figure*}

A deeper analysis shows that asymmetry in \faal{} and \specnorm{} stems from misclassifications between particular class pairs, where adversarial examples from one class are markedly more likely to be classified as the other class than vice versa.
For instance, in the case of the \pubfigsiblings{} dataset (\figref{fig:symmetry-all}), \faal{} and \specnorm{} are particularly asymmetric for classes 1 and 2 (Dakota and Elle Fanning, two sisters), 6 and 8 (Jennifer Lopez and Reese Witherspoon), and 6 and 9 (Jennifer Lopez and Beyonc\'e). 
By contrast, \defense{} yields higher symmetry, including for challenging classes where \faal{} and \specnorm{} struggle.
Said differently, with \defense{}, pairwise confusions are more symmetric and no class disproportionately ``loses'' to another.
Similar trends are observed for the CIFAR datasets (\tabref{tab:cifar10-100-sym} in \appref{app:cifar-results}) and targeted attacks \ifArxiv (\tabref{tab:targeted-symmetry} and \figref{fig:pubfig_siblings_tgt_attack_heatmaps} 
in \appref{app:tgtd_attack}).
\fi

\begin{table*}[t]
\centering
\small
\setlength{\tabcolsep}{5pt}
\renewcommand{\arraystretch}{1.1}
\begin{tabular}{@{}l c c c c c c@{}}
\toprule
\multirow{2}{*}{\textbf{Method}} 
& \multicolumn{2}{c}{\textbf{\pubfig}} 
& \multicolumn{2}{c}{\textbf{\pubfigsiblings}} 
& \multicolumn{2}{c}{\textbf{\pubfigvit}} \\
\cmidrule(lr){2-3}\cmidrule(lr){4-5}\cmidrule(lr){6-7}
& \textbf{Asym. Gap}~$\downarrow$ & \textbf{Sym. Loss}~$\downarrow$ 
& \textbf{Asym. Gap}~$\downarrow$ & \textbf{Sym. Loss}~$\downarrow$ 
& \textbf{Asym. Gap}~$\downarrow$ & \textbf{Sym. Loss}~$\downarrow$ \\
\midrule
Adv. Train & 0.5491 & 2.1812 & 0.3781 & 1.2967 & 0.3052 & 1.4260 \\
\faal                & 0.2728 & 1.0803 & 0.8159 & 1.8245 & 0.4210 & 1.7487 \\
\specnorm            & 0.1864 & 0.4854 & 0.6641 & 1.2933 & 0.2571 & 1.4029 \\
\textbf{Sy-FAR}      & \textbf{0.1704} & \textbf{0.4162} & \textbf{0.2320} & \textbf{0.7214} & \textbf{0.2386} & \textbf{1.3159} \\
\bottomrule
\end{tabular}
\caption{
The impact of different training approaches on symmetry, as captured by the Max Asymmetry Gap and the Symmetry Loss. The results are reported for three setups (\pubfig, \pubfigsiblings, and \pubfigvit), using the untargeted eyeglass attack to generate adversarial examples, and are averaged across ten runs.}
\label{tab:sym_untgtd}
\end{table*}

\subsection{Source-Class Robustness and Fairness}
\label{sec:source_class}
\label{sec:experiment_results:source}

We next explore the effect of optimizing symmetry as part of \defense{} on fair source-class adversarial robustness as well as models' overall accuracy and robustness. 
Specifically, we focus on benign accuracy, robust accuracy, worst-class robust accuracy, and robustness disparities between classes, as these metrics are the standard metrics in the literature on fair adversarial robustness ~\cite{xu2021robust,sun2023improving,zhang2024towards,jin2025enhancing}.

\tabref{tab:src_untgtd} presents the results for the face-recognition datasets and models. 
It can be seen that \defense{} preserves or even slightly improves benign accuracy compared to baseline approaches. 
In terms of robust accuracy, \defense{} consistently surpasses prior methods, with the most striking improvement on the challenging \pubfigsiblings{} setting: robustness improves by 6.5\% over the strongest baseline (\specnorm{}) and 10.7\% over \faal{}. 
This demonstrates that symmetry regularization remains effective even when visually similar identities exacerbate misclassifications.  

For worst-class accuracy and the fairness gap, the gains are even more dramatic. 
On \pubfigsiblings{}, compared to \faal{} and \specnorm{}, \defense{} improves the minimum robust accuracy by a factor of 62.6$\times$ and 3.5$\times$, respectively, while narrowing the disparity gap by 51.3\% and 40.9\%. 

\begin{figure*}[t!]
  \centering
  \begin{subfigure}{0.68\columnwidth}
    \centering
    \includegraphics[width=\linewidth]{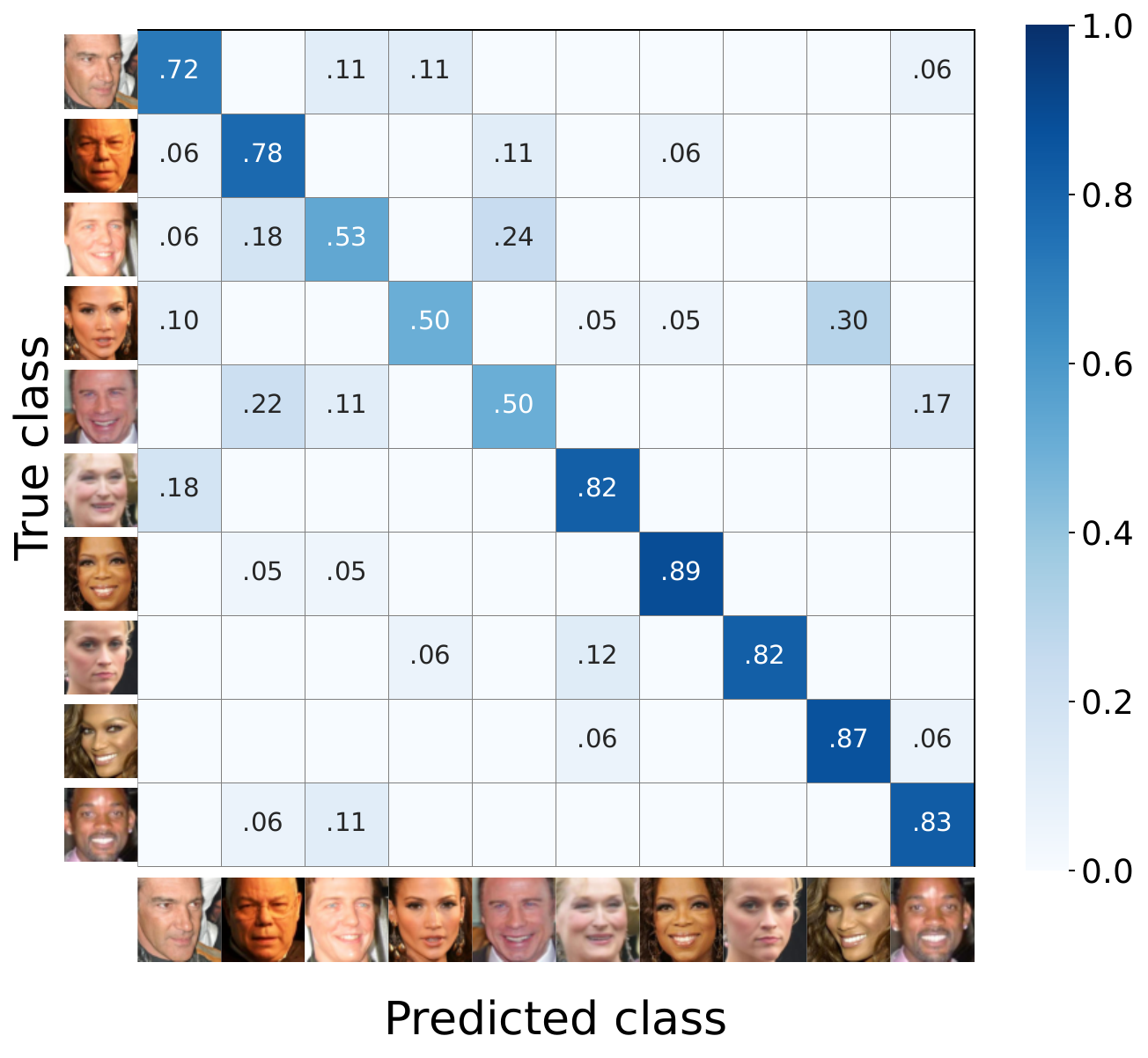}
    \caption{\faal}
  \end{subfigure}
  \hfill
  \begin{subfigure}{0.68\columnwidth}
    \centering
    \includegraphics[width=\linewidth]{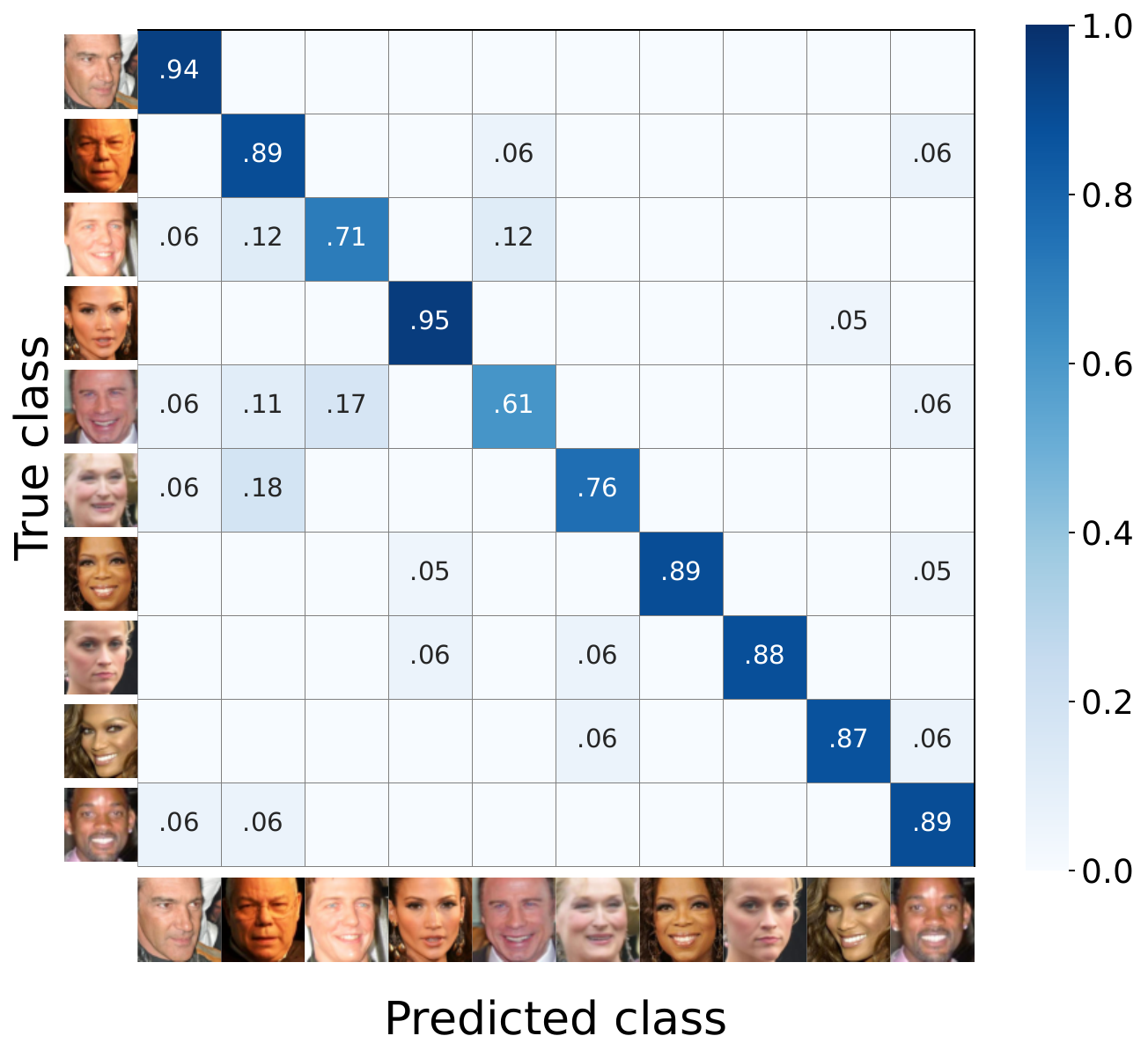}
    \caption{\specnorm}
  \end{subfigure}
  \hfill
  \begin{subfigure}{0.68\columnwidth}
    \centering
    \includegraphics[width=\linewidth]{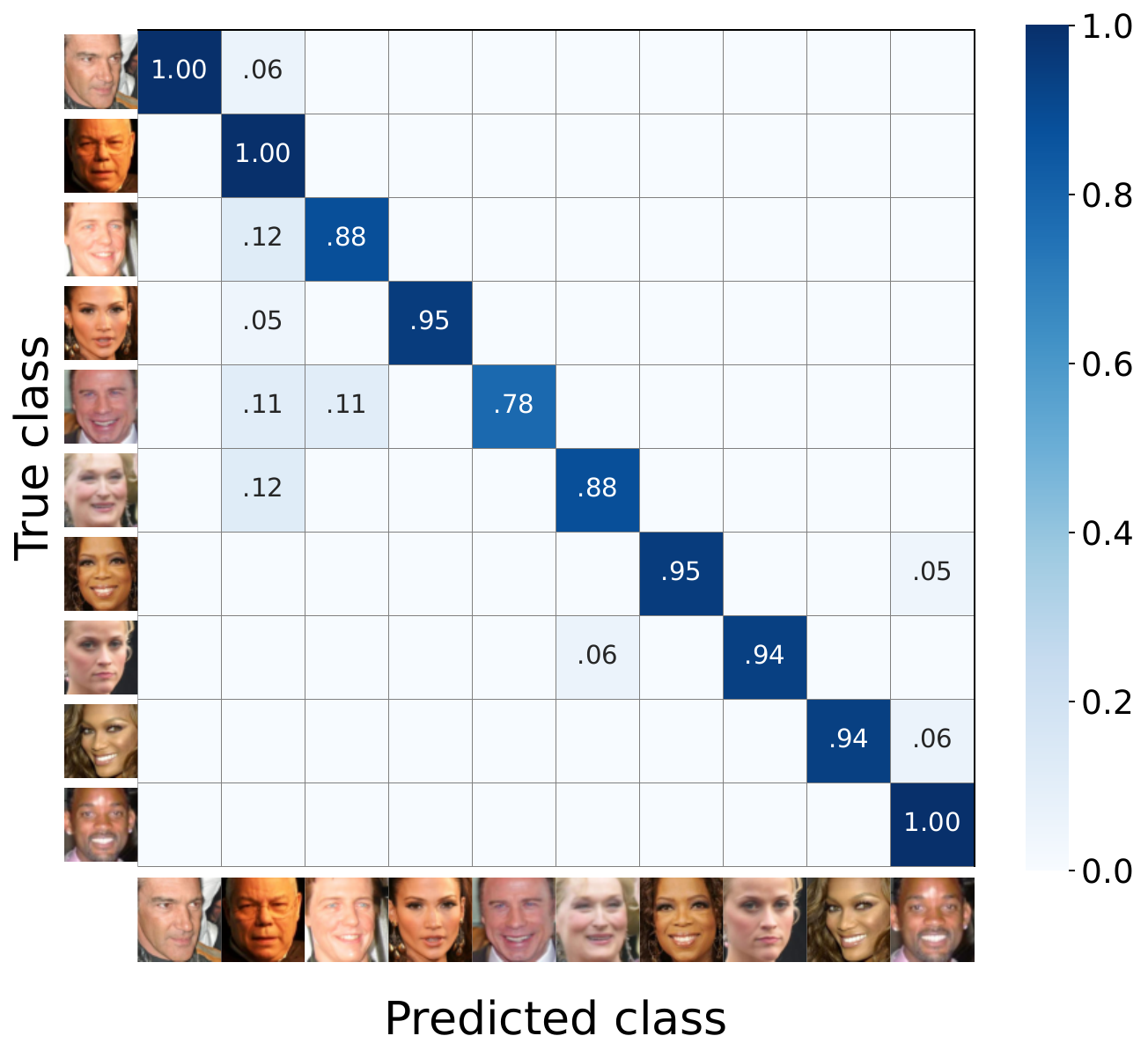}
    \caption{\defense{}}
  \end{subfigure}
    \caption{Confusion matrices for different methods on adversarial examples produced with untargeted eyeglass attack against the \pubfig{} setup. Diagonals indicate source-class robust accuracy; off-diagonals are misclassifications. We show representative heatmaps from individual randomly selected runs out of the ten repetitions.}
    \label{fig:mix_sibs_heatmaps}
\end{figure*}

Overall, these results show that explicitly enforcing symmetry yields both robustness and fairness gains even in the most challenging scenarios, while preserving high clean accuracy---a combination unmatched by established baselines.
As shown in Fig.~\ref{fig:mix_sibs_heatmaps}, \defense{} yields the strongest diagonals, reflecting the highest source-class robust accuracy among all methods.
Comparable patterns are evident for the CIFAR models (\tabref{tab:cifar10-100-acc} in \appref{app:cifar-results}) and for targeted attacks \ifArxiv (\tabref{tab:targeted-source-fairness} and \figref{fig:pubfig_siblings_tgt_attack_heatmaps} in \appref{app:tgtd_attack}).
\fi

\begin{table*}[t!]
\centering
\small
\setlength{\tabcolsep}{5pt}  %
\renewcommand{\arraystretch}{1.1}
\begin{tabular}{@{}lrrrrrrrrrrrr@{}}
\toprule
\multirow{2}{*}{\textbf{Method}} 
& \multicolumn{4}{c}{\textbf{\pubfig}} 
& \multicolumn{4}{c}{\textbf{\pubfigsiblings}} 
& \multicolumn{4}{c}{\textbf{\pubfigvit}} \\
\cmidrule(lr){2-5}\cmidrule(lr){6-9}\cmidrule(lr){10-13}
& \textbf{Benign}~$\uparrow$ & \textbf{Robust}~$\uparrow$ & \textbf{Min}~$\uparrow$ & \textbf{Gap}~$\downarrow$
& \textbf{Benign}~$\uparrow$ & \textbf{Robust}~$\uparrow$ & \textbf{Min}~$\uparrow$ & \textbf{Gap}~$\downarrow$
& \textbf{Benign}~$\uparrow$ & \textbf{Robust}~$\uparrow$ & \textbf{Min}~$\uparrow$ & \textbf{Gap}~$\downarrow$ \\
\midrule
Adv.\ Train & 96.91 & 58.34 &  2.50 & 92.62 & 95.35 & 68.62 & 28.65 & 67.39 & 82.02 & 63.22 & 32.25 & 61.66 \\
\faal          & 96.91 & 71.27 & 40.16 & 53.02 & 88.19 & 70.65 &  0.78 & 99.22 & 81.41 & 60.06 & 21.97 & 70.80 \\
\specnorm      & 98.26 & 86.41 & 66.25 & 33.75 & 93.04 & 73.46 & 14.06 & 81.84 & 83.39 & 65.68 & 30.86 & 62.86 \\
\textbf{Sy-FAR}& \textbf{98.53} & \textbf{89.44} & \textbf{70.72} & \textbf{29.28}
               & \textbf{95.47} & \textbf{78.23} & \textbf{48.85} & \textbf{48.37}
               & \textbf{83.87} & \textbf{67.82} & \textbf{33.66} & \textbf{58.57} \\
\bottomrule
\end{tabular}
\caption{
The impact of different training approaches on accuracy and source-class fairness. 
The results are reported for three setups (\pubfig, \pubfigsiblings, and \pubfigvit), using the untargeted eyeglass attack to generate adversarial examples, and are averaged across ten runs.
We report benign accuracy (Benign$\uparrow$), robust accuracy (Robust$\uparrow$), worst-class accuracy (Min$\uparrow$), and the class-level robust accuracy gap (Gap$\downarrow$).}
\label{tab:src_untgtd}
\end{table*}

\subsection{Target-Class Fairness}
\label{sec:target_class}
\label{sec:res:target}
\label{sec:experiment_results:target}

If a target class is more likely to be selected as the output of misclassified adversarial examples, this class could be significantly disadvantaged compared to others. 
Particularly, in the context of face recognition, such a class (i.e., individual) may be at a disproportionally higher risk of impersonation %
than other classes. 
Indeed, we find, to our knowledge, for the first time, that certain classes are markedly more likely to be selected as the output of adversarial examples than others.
\figref{fig:mix_sibs_heatmaps} depicts this form of unfair target-class robustness. 
Specifically, it can be seen that, for models trained via \faal{} or \specnorm{} on \pubfig{}, certain classes are disproportionally more likely to be selected as the output of misclassified adversarial examples (especially, Hugh Grant, Colin Powell, and Will Smith).
This issue is less pronounced for \defense{}.

\tabref{tab:tgt_untgtd} reports Target-Class Fairness metrics, which evaluate %
how misclassifications are distributed across target classes. 
We consider the maximum target confusion share (MaxTgt$\downarrow$), the minimum target confusion share (MinTgt$\uparrow$), and the standard deviation across classes (Std$\downarrow$).

In terms of MaxTgt, \defense{} consistently achieves the lowest concentration of errors across all datasets. 
On the challenging \pubfigsiblings{} setup, \defense{} significantly lowers the worst-case error share to 0.2008, compared to 0.4543 for \specnorm{} and 0.5254 for \faal{}, reducing the dominant error sink by 55.8\% and 61.8\%, respectively.  
For MinTgt, \defense{} raises the smallest target share across all setups, ensuring that no identity is systematically neglected as a predicted class and the gap between the most vulnerable and least vulnerable target classes is small.
Finally, for Std, which measures dispersion of target shares, \defense{} consistently yields the lowest values. 
On \pubfigsiblings{}, for example, it reduces dispersion to 0.0505 compared to 0.1226 for \specnorm{} and 0.1480 for \faal{}, cutting skew by 58.8\% and 65.9\%, respectively. 
Overall, the results confirm that enforcing symmetry improves not only robustness and source-class fairness, but also target-class fairness---ensuring that no single class disproportionately carries the burden of errors, even in the most challenging setups.

In \figref{fig:mix_sibs_heatmaps}, \defense{} also produces the faintest off-diagonals across columns, indicating fewer concentrated misclassifications and improved target-class fairness.
Consistent trends are also confirmed for CIFAR (\tabref{tab:cifar10-100-tgt} in \appref{app:cifar-results}) and targeted attacks \ifArxiv (\tabref{tab:targeted-target-fairness} and \figref{fig:pubfig_siblings_tgt_attack_heatmaps} in \appref{app:tgtd_attack}).
\fi

\begin{table*}[t]
\centering
\small
\setlength{\tabcolsep}{5pt}
\renewcommand{\arraystretch}{1.1}
\begin{tabular}{@{}lrrrrrrrrr@{}}
\toprule
\multirow{2}{*}{\textbf{Method}} 
& \multicolumn{3}{c}{\textbf{\pubfig}} 
& \multicolumn{3}{c}{\textbf{\pubfigsiblings}} 
& \multicolumn{3}{c}{\textbf{\pubfigvit}} \\
\cmidrule(lr){2-4}\cmidrule(lr){5-7}\cmidrule(lr){8-10}
& \textbf{MinTgt}~$\uparrow$ & \textbf{MaxTgt}~$\downarrow$ & \textbf{Std}~$\downarrow$
& \textbf{MinTgt}~$\uparrow$ & \textbf{MaxTgt}~$\downarrow$ & \textbf{Std}~$\downarrow$
& \textbf{MinTgt}~$\uparrow$ & \textbf{MaxTgt}~$\downarrow$ & \textbf{Std}~$\downarrow$ \\
\midrule
Adv.\ Training & 0.0000 & 0.3174 & 0.0983 & 0.0276 & 0.2353 & 0.0631 & 0.0162 & 0.2291 & 0.0700 \\
\faal          & 0.0048 & 0.3145 & 0.0880 & 0.0048 & 0.5254 & 0.1480 & 0.0001 & 0.2141 & 0.0725 \\
\specnorm      & 0.0099 & 0.3120 & 0.0886 & 0.0141 & 0.4543 & 0.1226 & 0.0146 & 0.1805 & 0.0601 \\
\textbf{Sy-FAR}& \textbf{0.0103} & \textbf{0.2563} & \textbf{0.0796} 
               & \textbf{0.0339} & \textbf{0.2008} & \textbf{0.0505} 
               & \textbf{0.0224} & \textbf{0.1709} & \textbf{0.0591} \\
\bottomrule
\end{tabular}
\caption{
The impact of different training approaches on target-class fairness. 
The results are reported for three setups (\pubfig, \pubfigsiblings, and \pubfigvit), using the untargeted eyeglass attack to generate adversarial examples, and are averaged across ten runs.
We report the minimum and maximum normalized misclassification into each target class (MinTgt$\uparrow$ and MaxTgt$\downarrow$, respectively) and the standard deviation (Std) across target classes.}
\label{tab:tgt_untgtd}
\end{table*}

\subsection{Run-Time Comparison}
\label{sec:runtime}
\label{sec:experiment_results:time}

\begin{figure}[t!]
    \centering   \includegraphics[width=0.77\linewidth]{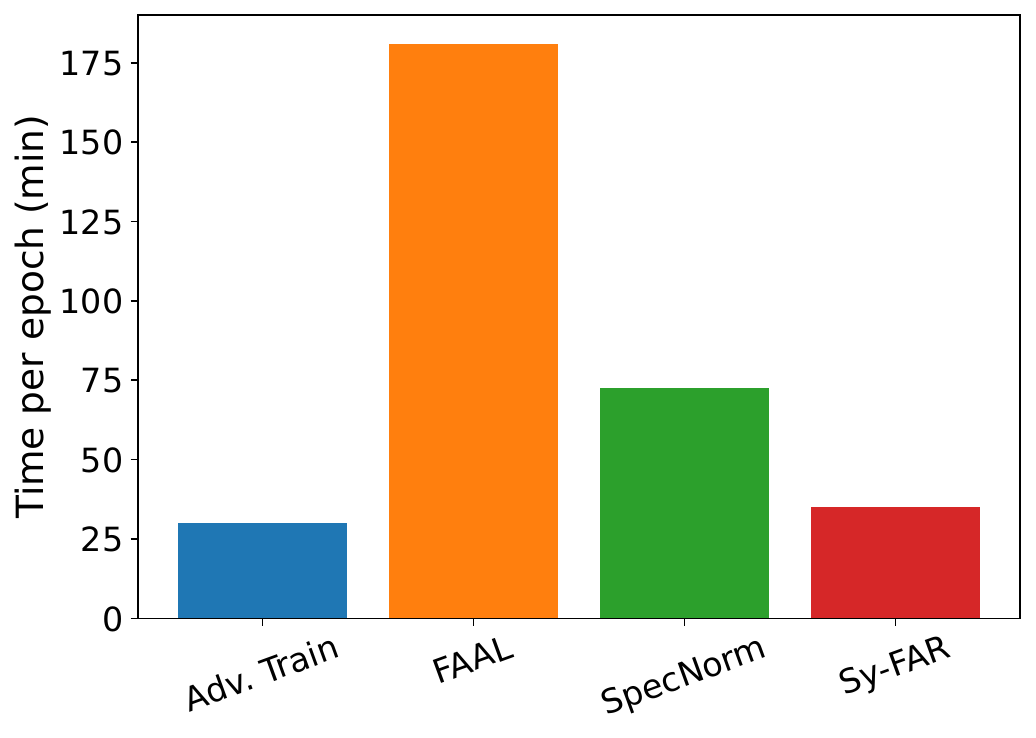}
    \caption{
    Training time per epoch across different methods. 
    }
    \label{fig:runtime-comparison}
\end{figure}

We compare the computational cost of \defense{} against the baseline methods on the \pubfig{} dataset using a VGG16 backbone. 
Experiments were conducted on a cluster node equipped with a single NVIDIA TITAN Xp GPU (12 GB), and the reported times are averaged across ten independent runs.
\figref{fig:runtime-comparison} shows that \defense{} is substantially more efficient than \faal{} and \specnorm{}.
By contrast, \faal{} is nearly five times slower than \defense{} because of its bi-level optimization procedure that repeatedly solves inner adversarial updates for fairness-aware weighting. 
\specnorm{} also incurs a significant slowdown---more than a factor of two---stemming from the need to compute the spectral norm of %
confusion matrix via singular value decomposition once per epoch. 
By comparison, standard adversarial training serves as the baseline cost (30 minutes/epoch), and \defense{} adds only a negligible overhead on top of it, since the $O(K^2)$ operations from the confusion matrix (\secref{sec:tech:alg}) are minor compared to the far larger cost of weight updates and adversarial example generation. 
Additional run-time analysis is included in
\ifArxiv
    \appref{app:scalability}.
\fi

\vspace{-5pt}

\subsection{Stability Across Runs}
\label{sec:stability}
\label{sec:experiment_results:stability}

Beyond average performance, a key requirement for defenses is \emph{stability}: 
producing consistently strong results across different 
runs. 
An unstable training method may require multiple attempts or fail to converge reliably under practical time and computational constraints.  
Indeed, we find that certain approaches are markedly more stable than others.
We measure stability by examining the variance in robust accuracy, worst class accuracy and accuracy gap across ten independent runs. 
The results (shown in Figs.~\ref{fig:stability-bar1}--\ref{fig:stability-bar3} in \appref{app:stability}) demonstrate that \defense{} achieves superior stability (i.e., lower variance) compared to alternative approaches not only in terms of robust accuracy, but also in terms of fair adversarial robustness.
One possible explanation is that \faal{} relies on bilevel optimization for fairness-aware weighting, which introduces additional non-convexity and sensitivity to initialization, while \specnorm{} applies global spectral constraints that can interact unpredictably with adversarial updates. 
By comparison, \defense{} introduces a symmetry regularization term that, we posit, acts as a smoother objective, %
reducing run-to-run variance without adding significant optimization instability.

\section{Fairness for Subgroup}
\label{sec:fairness_subgrp}
Ensuring fairness across all subgroups is a central challenge in \ml{}. 
As Rothblum and Yona~\cite{rothblum2018probably} point out, the number of potential subgroups in a classification task grows exponentially with the number of classes $K$. 
Specifically, the number of possible partitions of $K$ classes is given by the Bell number, which already exceeds 100{,}000 for $K$=10. 
Even if we restrict attention to fairness between pairs of disjoint subgroups, the combinatorial explosion remains substantial---there are over 28{,}000 such pairs when $K$=10---rendering exhaustive subgroup fairness enforcement computationally infeasible.

To address this, \defense sidesteps the need to explicitly define or enforce fairness for all possible subgroups. Instead, \defense{} leverages a simple yet effective regularizer that promotes pairwise symmetry in the confusion matrix: it directly penalizes asymmetries in the likelihood of misclassifying class $i$ as $j$ vs.\ $j$ as $i$. Consequently, as we now show, by minimizing directional imbalances across all class pairs, \defense{} promotes symmetry not only at the individual class level but also across any subgroup composed of these classes. While symmetry does not explicitly enforce fairness, our experiments demonstrate that encouraging such symmetry consistently leads to improved fairness outcomes at the subgroup level.
This property is powerful: when symmetry holds between all class pairs, it induces symmetry over arbitrary partitions and combinations of these classes. As a result, and as a by product of the regularization, \defense{}'s symmetry-based regularization enhances \emph{subgroup fairness}, without requiring explicit group labels, domain knowledge, or combinatorial enumeration.

\subsection{Defining (Sub)group Symmetry}
To define subgroup symmetry in classification, we build on the soft confusion matrix defined in \secref{sec:technical_approach}, where each entry ${C}_{ij}$ denotes the expected probability of predicting class $j$ given true class $i$. 
Recall that the matrix is constructed from softmax probabilities over adversarial examples and each row sums to one.
Let ${P}=\{G_1,\dots,G_m\}$ be a partition of the $K$ classes into $m$ disjoint subgroups, with $G_i\subseteq\{1,\dots,K\}$ and $G_i\cap G_j=\emptyset$ for $i\neq j$. 
Each subgroup $G_i$ may represent a demographic attribute (e.g., gender), a shared %
feature (e.g., similar appearance), or any meaningful grouping not necessarily known \emph{a priori} during training.
Using this notation, we now define the notion of symmetry at the (sub)group-level.

\begin{definition}[Subgroup Misclassification Rate]
Given a partition ${P} = \{G_1,\dots,G_m\}$ of the $K$ classes, the 
\emph{subgroup misclassification rate} from $G_a$ to $G_b$ is defined as
\[
\widehat{{C}}_{G_a G_b} := \frac{1}{n_a n_b}\sum_{i \in G_a} \sum_{j \in G_b} {C}_{ij}.
\]
where $n_a = |G_a|$ and $n_b = |G_b|$ are the subgroup sizes. 
\end{definition}
This quantity captures the average probability of misclassifying a sample from subgroup $G_a$ into any class in subgroup $G_b$, adjusted for subgroup sizes.

\begin{definition}[Subgroup Symmetry]
Two subgroups $G_a$ and $G_b$ are said to be \emph{symmetric} if their misclassification rates are approximately equal:
\[
\widehat{{C}}_{G_a G_b} \approx \widehat{{C}}_{G_b G_a}.
\]
\end{definition}
Subgroup symmetry ensures that errors between $G_a$ and $G_b$ are not directionally biased.

\subsection{From Individual to Subgroup Symmetry}
\label{sec:subgroup:thoery}

\defense{} promotes a novel form of fairness by enforcing symmetry in the confusion matrix, requiring that the probability of misclassifying class $i$ as $j$ matches the reverse direction, i.e., ${C}_{ij} = {C}_{ji}$ for all $i \neq j$. This criterion, which we refer to as \emph{class-level symmetry}, is both intuitive and efficient to compute.  
To ensure fairness at the \emph{group level}, however, we must show that the same principle extends to partitions of the label space. Formally, given any partition ${P} = \{G_1, \dots, G_m\}$ of the $K$ classes, we require that any two subgroups satisfy subgroup symmetry:
\[
\widehat{{C}}_{G_a G_b} \approx \widehat{{C}}_{G_b G_a}, \quad \forall\, G_a, G_b \in {P}, \; a \neq b.
\]
At first glance, enforcing such subgroup symmetry appears to demand evaluating an exponential number of subgroup pairs, which is computationally prohibitive. Remarkably, we prove that our class-level symmetry regularizer is sufficient: enforcing ${C}_{ij} = {C}_{ji}$ at the class level \emph{implies subgroup symmetry fairness} across all partitions and vice versa.
The key insight is that both class-level and subgroup-level symmetry arise from the same confusion matrix. Symmetry in individual off-diagonal entries naturally aggregates into symmetric relationships between groups. We now formally state and prove the equivalence between class- and subgroup-level symmetry.

\begin{theorem}
Let ${C} \in \mathbb{R}_{\ge 0}^{K \times K}$ be a normalized confusion matrix. 
Let ${P} = \{G_1, G_2, \dots, G_m\}$ be a partition of $\{1,\dots,K\}$ into disjoint subgroups, %
then:
\begin{enumerate}
    \item If ${C}$ is symmetric, i.e., (${C}_{ij} = {C}_{ji}$ for all $i \neq j$), then subgroup symmetry holds: $\widehat{{C}}_{G_a G_b} = \widehat{{C}}_{G_b G_a}$ for all $G_a \neq G_b$ in any partition ${P}$.
    \item Conversely, if subgroup symmetry holds for \emph{all} partitions ${P}$, then ${C}$ must be %
    symmetric.
\end{enumerate}
Said concisely, ${C}$ is symmetric if and only if $\widehat{{C}}$ is symmetric for all subgroup partitions.
\end{theorem}

\begin{proof}
\textbf{(1) Class symmetry $\Rightarrow$ Symmetry for subgroups.}
Assume $C_{ij} = C_{ji}$ for all $i \neq j$, and consider any two distinct subgroups $G_a, G_b$. By definition,
\\
\[
\widehat{C}_{G_a G_b} \;=\; \frac{1}{n_a n_b}\sum_{i \in G_a} \sum_{j \in G_b} C_{ij}
\;=\; \frac{1}{n_a n_b}\sum_{i \in G_a} \sum_{j \in G_b} C_{ji}.
\]

Switching the order of summation (finite sums) gives
\[
\widehat{C}_{G_a G_b}
\;=\; \frac{1}{n_a n_b}\sum_{j \in G_b} \sum_{i \in G_a} C_{ji}
\;=\; \widehat{C}_{G_b G_a}.
\]
Thus $\widehat{C}_{G_a G_b} = \widehat{C}_{G_b G_a}$ for all $a \neq b$.

\medskip
\noindent
\textbf{(2) Symmetry for all subgroup partitions $\Rightarrow$ Class Symmetry.}
Suppose $\widehat{C}_{G_a G_b} = \widehat{C}_{G_b G_a}$ holds for all partitions.
For two classes $i \ne j$, examine the following
three-part partition:
\[
{G} = \{G_1 = \{i\},\; G_2 = \{j\},\; G_3 = \{1,\dots,K\}\setminus\{i,j\}\}.
\]
We get that
\[
\widehat{C}_{G_1 G_2} = \frac{1}{1\cdot 1}\sum_{p \in G_1}\sum_{q \in G_2} C_{pq} = C_{ij}, 
\]
and
\[
\widehat{C}_{G_2 G_1} = \frac{1}{1\cdot 1}\sum_{p \in G_2}\sum_{q \in G_1} C_{pq} = C_{ji}.
\]
(Notice that $n_1 = n_2 = 1$.)
By subgroup symmetry, $\widehat{C}_{G_1 G_2}=\widehat{C}_{G_2 G_1}$, hence $C_{ij}=C_{ji}$. 
As this equality holds for any pair of classes $i\neq j$, $C$ is class symmetric.
\end{proof}

\subsection{Empirically Assessing Subgroup Fairness}
\label{sec:subgroup:res}

In \secref{sec:experiment_results}, we empirically showed that \defense{} promotes fairness for individuals under evasion attacks;
in particular, \defense{} increases symmetry as well as source- and target-class fair adversarial robustness.
Complementing the theoretical result from \secref{sec:subgroup:thoery}, we now empirically demonstrate that optimizing for symmetry at the level of individual classes also improves symmetry, and as a by-product, source- and target-group fair adversarial robustness, for subgroups. 
Notably, these improvements are attained in a manner completely agnostic to the subgroup definitions, as \defense{} does not receive any subgroup information as part of its input.

\paragraph{Datasets, Models, and Groups.}
We evaluated all three face-recognition setups presented in \secref{sec:experiment_setup}.
To study subgroup-level fairness, we focus on two demographic attributes provided in the PubFig dataset~\cite{attribute_classifiers}: 
\emph{gender} (male vs.\ female) and \emph{ethnicity} (white vs.\ non-white). 
Class-to-subgroup assignments follow directly from the PubFig annotations.

\paragraph{Evaluation Metrics.}
To evaluate robustness and fairness at the level of subgroups, we  report robust accuracy for each subgroup separately. 
The differences between subgroups' robust accuracies inform us about the degree of unfair source-group robust accuracy.
Note that, because we consider two subgroups in each of the partitions, differences in unfair target-group robust accuracies are similar to the differences in source-group robust accuracy.

\begin{figure}[t!]
    \centering
    \includegraphics[width=0.8\linewidth]{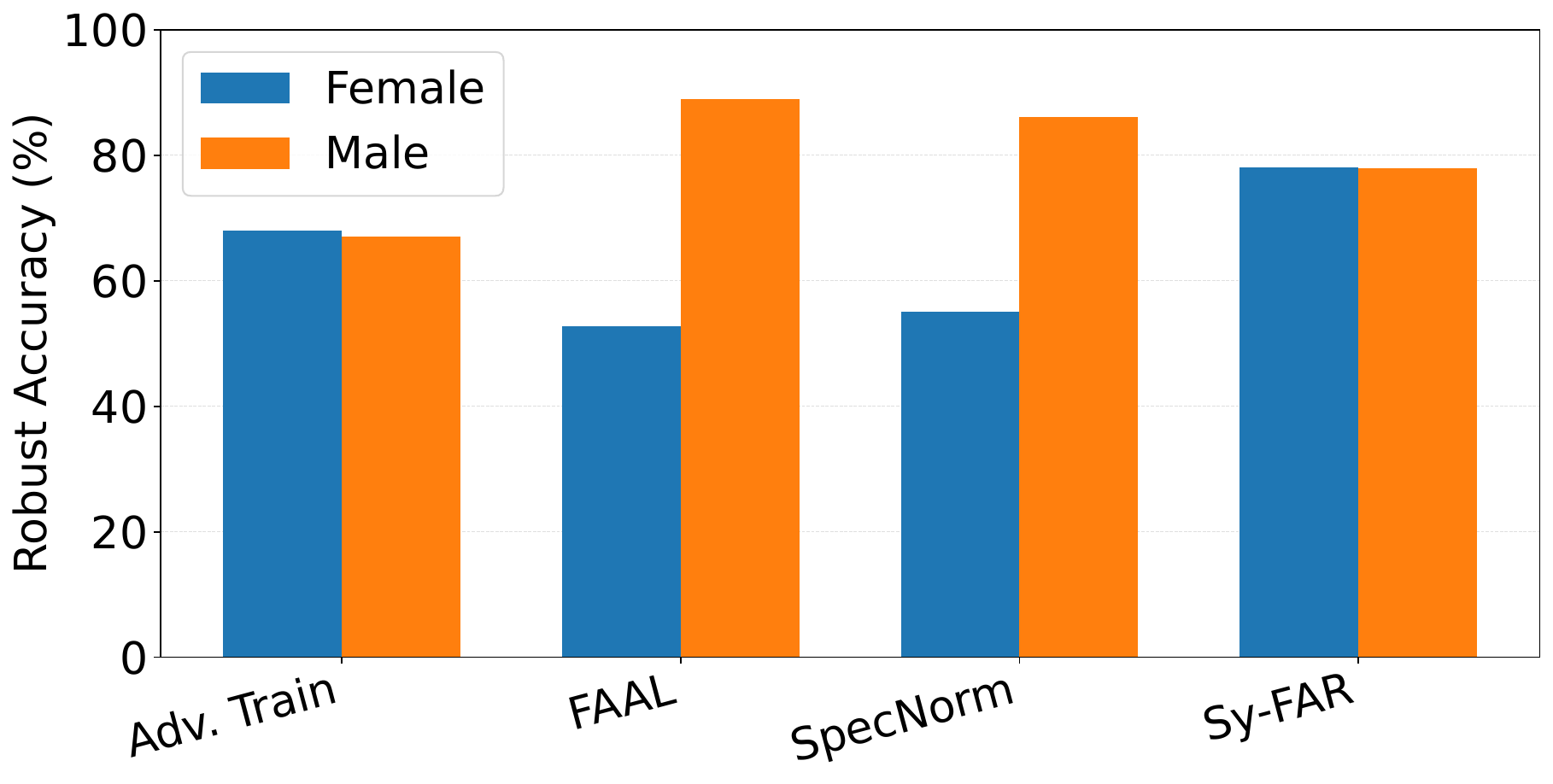}
    \caption{
    Subgroup robust accuracy by gender (male vs.\ female) on \pubfigsiblings. 
    }
    \label{fig:grouped_gender}
\end{figure}

\paragraph{Evaluation Results.}
\figref{fig:grouped_gender} reports subgroup robust accuracy for gender on the \pubfigsiblings{} setup
between males and females. It can be observed that adversarial training exhibits a small gap between males and females (0.08), but the overall robust accuracy is low. 
\faal{} and \specnorm{} achieve higher overall robust accuracy than adversarial training; however, males attain significantly higher robust accuracy, with the gender gap reaching $0.36$ for \faal{} and $0.32$ for \specnorm{}.
By contrast, \defense{} not only achieves the highest overall robust accuracy, but the gap between males and females is the lowest (0.04), with females receiving higher robust accuracy than males.
\ifArxiv The results for datasets other than \pubfigsiblings{} show identical trends (\tabref{tab:gaps-gender} in \appref{app:subgroup}), albeit \defense{} fares best in the challenging \pubfigsiblings{} setting.
The results for the ethnicity partition (\figref{fig:grouped_color} and \tabref{tab:gaps-ethnicity} in \appref{app:subgroup}) show similar trends as those for the gender partition, with similar levels of robust accuracy across gender and ethnicity groups.
\fi
Note that on all settings and between all subgroups, the gaps in benign accuracy between subgroups is small ($\le$0.04).
Overall, these results confirm that symmetry-based regularization not only encourages class-level fair adversarial robustness, but also propagates to subgroup-level fair adversarial robustness, substantially reducing demographic disparities under attacks.

\section{Limitations and Future Work}
\label{sec:limitations}

While \defense{} achieves strong and consistent improvements in fair adversarial robustness, several limitations remain, opening promising directions for future research. First, our experiments were conducted under controlled and balanced conditions to isolate the core fairness–robustness challenge. We show that this problem is already difficult in balanced datasets and becomes considerably harder when bias is introduced, as seen in the \pubfigsiblings{} setup. Extending \defense{} to highly imbalanced or noisy datasets, where labels may be uncertain or ambiguous, is an important next step toward deployment in real-world systems.

Second, the theoretical proof that class-level symmetry implies subgroup-level fairness assumes a finite and correctly labeled class space. In open-world or continuous-class settings—where new or unseen classes may appear—these guarantees may not strictly hold. Future work should explore how to generalize the concept of symmetry to continuous or open-set domains, possibly through embedding-based or probabilistic formulations.

Third, similar to prior work, this work focuses on classification and vision tasks, further extending them to security-critical applications such as face recognition. We believe that the underlying idea of symmetry can also extend to other modalities (e.g., natural language or audio). However, these domains may require domain-specific definitions of symmetric behavior, as similarity definitions and misclassification patterns may differ fundamentally from those in vision.

Finally, extending Sy-FAR to generative and multimodal models will likely require new formalizations of symmetry and fairness that account for continuous output spaces. Investigating these directions constitutes an interesting avenue for future work.

\section{Conclusion}
\label{sec:conclusion}
\label{sec:conclude}

This work introduced \defense{}, an approach for enhancing fair source- and target-class adversarial robustness through symmetry.
\defense{} is based on the intuitive insight that class resemblance is usually a symmetric relationship, i.e.,  misclassifications of adversarial examples between classes should be symmetric.
We extensively evaluate \defense{}, comparing it leading approaches~\cite{jin2025enhancing, zhang2024towards}, on fairness- and security-critical face-recognition tasks against realistic eyeglass attacks as well as standard object-classification benchmarks commonly used in the literature.
Our results, among others, demonstrate substantial improvements in fair source-class adversarial robustness while benign and robust accuracy are preserved or improved; 
identify unfair target-class robustness as a risk that \defense{} helps ameliorate; 
extend to group-based fairness where we show theoretically and empirically how symmetry advances fairness for arbitrary sub-groups in a computationally efficient manner.
Beyond fairness, \defense{} offers strong practical advantages: 
It incurs only negligible overhead relative to standard adversarial training, in sharp contrast to established methods, and it exhibits lower variance across independent runs, highlighting its stability and reproducibility.  

Taken together, our work shows that enforcing symmetry is an effective and principled way to mitigate biased error concentration while preserving robustness and scalability. 
We view \defense{} as a step toward fair and reliable adversarial training for safety-critical \ml{}-based systems, and anticipate that the idea of symmetry regularization can be extended to broader domains and threat models in future work.

\section*{Ethical Considerations}

Our work proposes \defense{}, a training method aimed at improving fairness and adversarial robustness of \ml{} models, with a particular focus on face-recognition tasks. Below, we discuss the relevant stakeholders, the ethical implications of our research process and publication, and the measures we took to mitigate potential negative impacts.

\paragraph{Stakeholders.}
The stakeholders relevant to this work include:
\begin{itemize}[nosep,noitemsep]
    \item individuals whose images appear in publicly available datasets used in our experiments;
    \item end users of ML-based security systems that may adopt fairness- and robustness-enhancing methods;
    \item organizations and practitioners developing, deploying, or auditing such systems;
    \item members of society at large, particularly populations that may be disproportionately affected by face recognition or automated decision-making; and
    \item the research community, including researchers building on our methods or findings.
\end{itemize}

\paragraph{Research Process and Experimental Scope.}
We evaluate face- and object-recognition models using only existing, publicly available datasets. Importantly, no human subjects or deployed systems were involved in our experiments. We did not collect new biometric data, scrape online platforms, or attempt to identify or track individuals beyond the labels already present in the datasets. We used each public dataset strictly in accordance with its published usage guidelines and terms, without augmenting it with any additional personal information.

All experiments are conducted offline, and we apply only known adversarial perturbation techniques introduced in prior work. We do not introduce new attack vectors that could increase risk to existing systems. As a result, our work does not expose new vulnerabilities or create additional attack capabilities beyond what is already documented in the adversarial ML literature. We do not experiment with live systems and do not violate any terms of service.

\paragraph{Positive and Negative Potential Impacts.}
On the positive side, Sy-FAR has the potential to improve fairness and promote more equal treatment across demographic groups in ML-based security systems. By encouraging more symmetric misclassification behavior and reducing worst-class performance gaps, our work can help identify and mitigate biases that disproportionately harm certain individuals or groups. Our analysis also contributes to academic understanding of fairness--robustness trade-offs, supporting the design and evaluation of fairer and more trustworthy ML systems.

At the same time, we acknowledge that improvements in robustness and fairness may have dual-use implications. More effective face-recognition systems could be misused in surveillance or monitoring applications that threaten privacy or civil liberties, particularly when deployed without transparency or regulation. More generally, technical improvements alone do not address broader social or political concerns regarding the appropriateness of such systems. Our goal is not to promote or legitimize such deployments. Rather, we aim to highlight (un)fairness and robustness issues in existing models and to inform the research community about potential trade-offs and mitigation strategies. We do not endorse the deployment of face-recognition or surveillance systems, particularly in high-risk or sensitive contexts.

\paragraph{Mitigations and Ethical Choices.}
To mitigate potential negative impacts, we make several deliberate choices in how we design and conduct our study. We restricted experiments to offline evaluations on public datasets, avoided live systems and real users, and used only known adversarial methods. We did not scrape data, violate terms of service, or introduce new attack techniques. Throughout the paper, we present Sy-FAR primarily as a mitigation approach for these issues during training, rather than as a justification for deploying face-recognition systems. These limitations and caveats are clearly stated to ensure responsible interpretation and use of our results.

\paragraph{Deception and Researcher Wellbeing.}
There were no user studies, no deception, and no involvement of outside participants. Consequently, there was no risk of harm to participants. The research did not involve exposure to disturbing content or conditions affecting researcher wellbeing, and standard academic safeguards were sufficient.

\paragraph{Reflection.}
We believe it is important to publish this work despite potential dual-use risks, as failing to study and report fairness, robustness issues, and possible mitigations may allow biased or unevenly robust systems to persist unexamined. However, we emphasize that technical solutions alone cannot resolve all ethical challenges surrounding biometric technologies. We hope this work informs both technical research and broader discussions on governance, regulation, and responsible use.

\section*{Open Science}
Our code repository\footnote{\url{https://zenodo.org/records/17901662} or \url{https://github.com/haneenn24/Sy-FAR-Symmetry-based-Fair-Adversarial-Robustness}.} includes all necessary artifacts to fully reproduce the results of the paper. A detailed \texttt{README} file provides step-by-step instructions on how to set up the environment, run training, and reproduce both our baseline adversarial training and the proposed fairness-aware methods (\faal{}~\cite{zhang2024towards}, \specnorm~\cite{jin2025enhancing}, and \defense{}). The instructions cover both clean training and adversarial training pipelines, as well as evaluation against benign and adversarial inputs (e.g., created with the eyeglass attack). 
To facilitate reproduction of results without training models from scratch, in addition to the source code, we provide pre-trained model checkpoints under reported settings trained with different robustness–fairness objectives.
All artifacts are documented to allow end-to-end reproduction of experiments, ensuring transparency and verifiability of our contributions.

\section*{Acknowledgments}
    This work has been supported in part
    by grant No.\ 2023641 from the United States-Israel Binational Science Foundation (BSF);
    by ISF grant no.\ 1807/23;
    by Len Blavatnik and the Blavatnik Family foundation;
    by a Maof prize for outstanding young scientists;
    by the Ministry of Energy, Israel (grant number 224-11-026);
    by the Ministry of Innovation, Science \& Technology, Israel (grant number 1001702114);
    by the Stellar Development Foundation;
    and by a grant from the Tel Aviv University Center for AI and Data Science (TAD).

\bibliographystyle{plain}
\bibliography{main}
\appendix
\section{Evaluation on CIFAR-10 and CIFAR-100}
\label{app:cifar-results}

While our primary focus is on face recognition, where fairness disparities pose immediate security and social risks, we also evaluate \defense{} on the widely used CIFAR-10 and CIFAR-100 benchmarks with the PreAct-ResNet18 model. 
These datasets and model are standard in the adversarial robustness and fairness literature~\cite{zhang2024towards,jin2025enhancing}, enabling direct comparison with prior methods. 
Since they lack real-world fairness challenges such as demographic balance or sibling similarity, we present the detailed results here in the Appendix, keeping the main text centered on security-critical face recognition scenarios.

\paragraph{Results Summary.}  
Across both CIFAR-10 and CIFAR-100, \defense{} achieves the strongest balance of robustness and fairness.  
\tabref{tab:cifar10-100-acc} shows that \defense{} improves robust accuracy over adversarial training and \faal, while matching or slightly exceeding \specnorm, particularly on worst-class accuracy.  
In terms of asymmetry (\tabref{tab:cifar10-100-sym}), \defense{} consistently reduces directional bias, achieving the lowest asymmetry gap and symmetry loss on both benchmarks.  
Finally, target-class fairness results in \tabref{tab:cifar10-100-tgt} confirm that \defense{} distributes misclassifications more evenly.  
Together, these results demonstrate that even in standard object classification benchmarks, \defense{} delivers robust improvements across all fairness perspectives while preserving competitive robustness.

\begin{table*}[t]
\centering
\small
\setlength{\tabcolsep}{6pt}
\renewcommand{\arraystretch}{1.1}
\begin{tabular}{@{}lrr@{\quad}lrr@{}}
\toprule
& \multicolumn{2}{c}{\textbf{CIFAR-10}} & & \multicolumn{2}{c}{\textbf{CIFAR-100}} \\
\cmidrule(lr){2-3}\cmidrule(lr){5-6}
\textbf{Method} 
& \textbf{Asym. Gap}~$\downarrow$ 
& \textbf{Sym. Loss}~$\downarrow$
& &
\textbf{Asym. Gap}~$\downarrow$ 
& \textbf{Sym. Loss}~$\downarrow$ \\
\midrule
Adv.\ Train   & 11.2 & 0.81 && 17.6 & 53.46 \\
\faal         &  9.8 & 1.17 && 15.9 & 38.80 \\
\specnorm     &  8.5 & 0.71 && 14.4 & 39.84 \\
\textbf{Sy-FAR} & \textbf{6.7} & \textbf{0.63} && \textbf{12.2} & \textbf{37.66} \\
\bottomrule
\end{tabular}
\caption{Symmetry on CIFAR-10/100 (PRN18):
The impact of different training approaches on symmetry, as captured by the Max Asymmetry Gap and the Symmetry Loss. 
Models are adversarially trained with PGD-\ensuremath{\ell_\infty} and evaluated with AutoAttack~\cite{Croce20Auto}. 
Results are averaged across ten runs. 
We report the Max Asymmetry Gap (Gap\ensuremath{\downarrow}) and the Symmetry Loss (Loss\ensuremath{\downarrow}).}
\label{tab:cifar10-100-sym}
\end{table*}

\begin{table*}[t]
\centering
\small
\setlength{\tabcolsep}{6pt}
\renewcommand{\arraystretch}{1.1}
\begin{tabular}{@{}lrrrr@{\quad}lrrrr@{}}
\toprule
& \multicolumn{4}{c}{\textbf{CIFAR-10}} & & \multicolumn{4}{c}{\textbf{CIFAR-100}} \\
\cmidrule(lr){2-5}\cmidrule(lr){7-10}
\textbf{Method} 
& \textbf{Benign}~$\uparrow$ 
& \textbf{Robust}~$\uparrow$ 
& \textbf{Min}~$\uparrow$
& \textbf{Gap}~$\downarrow$
& &
\textbf{Benign}~$\uparrow$ 
& \textbf{Robust}~$\uparrow$ 
& \textbf{Min}~$\uparrow$
& \textbf{Gap}~$\downarrow$ \\
\midrule
Adv.\ Train   & 82.07 & 47.12 & 12.90 & 34.22 && 54.70 & 22.15 & 1.00 & 21.15 \\
\faal         & 82.38 & 49.80 & 33.50 & 16.30 && 56.82 & 21.96 & 3.00 & 18.96 \\
\specnorm     & \textbf{83.51} & 49.94 & 34.94 & 15.00 && 57.36 & 25.17 & \textbf{4.00} & 21.17 \\
\textbf{Sy-FAR} & 82.02 & \textbf{52.60} & \textbf{35.31} & \textbf{17.29} && \textbf{57.60} & \textbf{25.77} & \textbf{4.00} & \textbf{21.77} \\
\bottomrule
\end{tabular}
\caption{
Accuracy and source-class fairness on CIFAR-10/100 (PRN18):
The impact of different training approaches on accuracy and fairness across true classes.
Models are adversarially trained with PGD-$\ell_\infty$ and evaluated with AutoAttack~\cite{Croce20Auto}.
Results are averaged across ten runs.
We report benign accuracy (Benign$\uparrow$), robust accuracy (Robust$\uparrow$), worst-class accuracy (Min$\uparrow$), and the class-level robust accuracy gap (Gap$\downarrow$).
}
\label{tab:cifar10-100-acc}
\end{table*}

\begin{table*}[t]
\centering
\small
\setlength{\tabcolsep}{6pt}
\renewcommand{\arraystretch}{1.1}
\begin{tabular}{@{}lrrr@{\quad}lrrr@{}}
\toprule
& \multicolumn{3}{c}{\textbf{CIFAR-10}} & & \multicolumn{3}{c}{\textbf{CIFAR-100}} \\
\cmidrule(lr){2-4}\cmidrule(lr){6-8}
\textbf{Method} 
& \textbf{MinTgt}~$\uparrow$ 
& \textbf{MaxTgt}~$\downarrow$ 
& \textbf{Std}~$\downarrow$
& &
\textbf{MinTgt}~$\uparrow$ 
& \textbf{MaxTgt}~$\downarrow$ 
& \textbf{Std}~$\downarrow$ \\
\midrule
Adv.\ Train   & 0.03 & 0.29 & 0.085 && 0.004 & 0.15 & 0.041 \\
\faal         & 0.04 & 0.24 & 0.071 && 0.006 & 0.13 & 0.036 \\
\specnorm     & 0.05 & 0.22 & 0.066 && 0.007 & 0.11 & 0.033 \\
\textbf{Sy-FAR} & \textbf{0.07} & \textbf{0.18} & \textbf{0.052} && \textbf{0.010} & \textbf{0.09} & \textbf{0.028} \\
\bottomrule
\end{tabular}
\caption{
Target-class fairness on CIFAR-10/100 (PRN18):
The impact of different training approaches on the distribution of misclassifications across predicted classes.
Models are adversarially trained with PGD-$\ell_\infty$ and evaluated with AutoAttack~\cite{Croce20Auto}.
Results are averaged across ten runs.
We report the minimum normalized misclassification into a target class (MinTgt$\uparrow$), the maximum normalized misclassification into a target class (MaxTgt$\downarrow$), and the standard deviation across classes (Std$\downarrow$).
}
\label{tab:cifar10-100-tgt}
\end{table*}
\section{Evaluation Under Targeted Eyeglass Attack}
\label{app:tgtd_attack}
In addition to the untargeted setting, we also evaluate all methods under the \emph{targeted eyeglass attack}, where adversarial perturbations are optimized to force misclassification specifically into a chosen (incorrect) target identity.  
We run the same experimental protocol across all three datasets (\pubfig, \pubfigsiblings, and \pubfigvit). 
Consistent with the untargeted results, \defense achieves the strongest improvements across all metrics, showing the lowest asymmetry and fairness disparities while maintaining high robustness. 
These findings confirm that symmetry regularization is effective not only against arbitrary misclassifications but also in resisting targeted adversarial attempts that deliberately push predictions toward specific identities.
\ifArxiv

\paragraph{Results Summary.}
Under the \emph{targeted} eyeglass attack, \defense consistently improves fairness across all dimensions. \tabref{tab:targeted-symmetry} shows that it achieves the best symmetry loss, reflecting reduced directional misclassification, a trend also visible in Fig.~\ref{fig:pubfig_siblings_tgt_attack_heatmaps}. 
In these heatmaps, each cell represents the success rate of forcing an input from source class $i$ to be misclassified as target class $j$. 
The diagonal is excluded, and darker off-diagonal regions correspond to higher attack success rates. 
Compared to all baselines, \defense{} exhibits the lowest off-diagonal intensities, indicating stronger robustness against targeted misclassifications. 

Moreover, as summarized in Tables~\ref{tab:targeted-source-fairness}--\ref{tab:targeted-target-fairness}, Sy-FAR strengthens fair source-class adversarial robustness, narrows per-class disparities, and distributes misclassifications more evenly across targets. Overall, these results demonstrate that Sy-FAR preserves robustness while advancing fairness even under targeted attacks.

\begin{table*}[t]
\centering
\small
\setlength{\tabcolsep}{6pt}
\renewcommand{\arraystretch}{1.1}
\begin{tabular}{@{}lrrrrrr@{}}
\toprule
\multirow{2}{*}{\textbf{Method}}
& \multicolumn{2}{c}{\textbf{\pubfig}}
& \multicolumn{2}{c}{\textbf{\pubfigsiblings}}
& \multicolumn{2}{c}{\textbf{\pubfigvit}} \\
\cmidrule(lr){2-3}\cmidrule(lr){4-5}\cmidrule(lr){6-7}
& \textbf{Asym.\ Gap}~$\downarrow$ & \textbf{Sym.\ Loss}~$\downarrow$
& \textbf{Asym.\ Gap}~$\downarrow$ & \textbf{Sym.\ Loss}~$\downarrow$
& \textbf{Asym.\ Gap}~$\downarrow$ & \textbf{Sym.\ Loss}~$\downarrow$ \\
\midrule
Adv.\ Training & 0.64 & 5.7633 & 0.32 & 2.0882 & 0.27 & 1.9421 \\
\faal          & 0.20 & 1.3151 & 0.42 & 2.2802 & 0.51 & 2.3176 \\
\specnorm      & 0.49 & 2.2102 & 0.68 & 1.8345 & 0.35 & 1.6044 \\
\textbf{Sy-FAR}& \textbf{0.18} & \textbf{0.9722} & \textbf{0.16} & \textbf{0.5578} & \textbf{0.19} & \textbf{0.7389} \\
\bottomrule
\end{tabular}
\caption{
The impact of different training approaches on symmetry, as captured by the Max Asymmetry Gap and the Symmetry Loss. The results are reported for three setups (\pubfig, \pubfigsiblings, and \pubfigvit), using the \emph{targeted eyeglass attack} to generate adversarial examples, and are averaged across ten runs.}
\label{tab:targeted-symmetry}
\end{table*}

\begin{table*}[t]
\centering
\small
\setlength{\tabcolsep}{6pt}
\renewcommand{\arraystretch}{1.1}
\begin{tabular}{@{}lrrrrrrrrr@{}}
\toprule
\multirow{2}{*}{\textbf{Method}}
& \multicolumn{3}{c}{\textbf{\pubfig}}
& \multicolumn{3}{c}{\textbf{\pubfigsiblings}}
& \multicolumn{3}{c}{\textbf{\pubfigvit}} \\
\cmidrule(lr){2-4}\cmidrule(lr){5-7}\cmidrule(lr){8-10}
& \textbf{Robust}~$\uparrow$ & \textbf{Min}~$\uparrow$ & \textbf{Gap}~$\downarrow$
& \textbf{Robust}~$\uparrow$ & \textbf{Min}~$\uparrow$ & \textbf{Gap}~$\downarrow$
& \textbf{Robust}~$\uparrow$ & \textbf{Min}~$\uparrow$ & \textbf{Gap}~$\downarrow$ \\
\midrule
Adv.\ Training & 99.8 & 99.0 & 1.0 & 98.0 & 95.0 & 5.0 & 99.7 & 96.0 & 4.0 \\
\faal          & 99.6 & 97.0 & 3.0 & 98.5 & 92.0 & 8.0 & 99.5 & 93.0 & 7.0 \\
\specnorm      & 99.4 & 94.0 & 6.0 & 98.6 & \textbf{94.0} & 6.0 & 99.2 & \textbf{98.0} & 2.0 \\
\textbf{Sy-FAR}& \textbf{100.0} & \textbf{100.0} & \textbf{0.0} & \textbf{98.7} & 93.0 & \textbf{1.0} & \textbf{99.9} & 95.0 & \textbf{1.0} \\
\bottomrule
\end{tabular}
\caption{
The impact of different training approaches on accuracy and source-class fairness.
The results are reported for three setups (\pubfig, \pubfigsiblings, and \pubfigvit), using the \emph{targeted eyeglass attack} to generate adversarial examples, and are averaged across ten runs.
We report benign accuracy (Benign$\uparrow$), robust accuracy Robust$\downarrow$, worst-class accuracy (Min$\uparrow$), and the class-level robust accuracy gap (Gap$\downarrow$).}
\label{tab:targeted-source-fairness}
\end{table*}

\begin{table*}[t]
\centering
\small
\setlength{\tabcolsep}{6pt}
\renewcommand{\arraystretch}{1.1}
\begin{tabular}{@{}lrrrrrrrrr@{}}
\toprule
\multirow{2}{*}{\textbf{Method}}
& \multicolumn{3}{c}{\textbf{\pubfig}}
& \multicolumn{3}{c}{\textbf{\pubfigsiblings}}
& \multicolumn{3}{c}{\textbf{\pubfigvit}} \\
\cmidrule(lr){2-4}\cmidrule(lr){5-7}\cmidrule(lr){8-10}
& \textbf{MinTgt}~$\uparrow$ & \textbf{MaxTgt}~$\downarrow$ & \textbf{Std}~$\downarrow$
& \textbf{MinTgt}~$\uparrow$ & \textbf{MaxTgt}~$\downarrow$ & \textbf{Std}~$\downarrow$
& \textbf{MinTgt}~$\uparrow$ & \textbf{MaxTgt}~$\downarrow$ & \textbf{Std}~$\downarrow$ \\
\midrule
Adv.\ Training & 0.0349 & 0.1481 & 0.0440 & 0.0147 & 0.2543 & 0.0700 & 0.0115 & 0.1527 & 0.0448 \\
\faal          & \textbf{0.0512} & 0.1491 & 0.0432 & 0.0350 & \textbf{0.2381} & 0.0360 & 0.0298 & 0.1689 & 0.0453 \\
\specnorm      & 0.0364 & 0.1842 & 0.0460 & \textbf{0.0360} & 0.3356 & 0.0817 & \textbf{0.0352} & 0.1763 & 0.0461 \\
\textbf{Sy-FAR}& 0.0350 & \textbf{0.1464} & \textbf{0.0430} & 0.0330 & 0.2959 & \textbf{0.0310} & 0.0331 & \textbf{0.1410} & \textbf{0.0427} \\
\bottomrule
\end{tabular}
\caption{
The impact of different training approaches on target-class fairness.
The results are reported for three setups (\pubfig, \pubfigsiblings, and \pubfigvit), using the \emph{targeted eyeglass attack} to generate adversarial examples, and are averaged across ten runs.
We report the minimum and maximum normalized misclassification into each target class (MinTgt$\uparrow$ and MaxTgt$\downarrow$, respectively) and the standard deviation (Std) across target classes.}
\label{tab:targeted-target-fairness}
\end{table*}

\begin{figure*}[t!]
  \centering
  \begin{subfigure}{0.68\columnwidth}
    \centering
    \includegraphics[width=\linewidth]{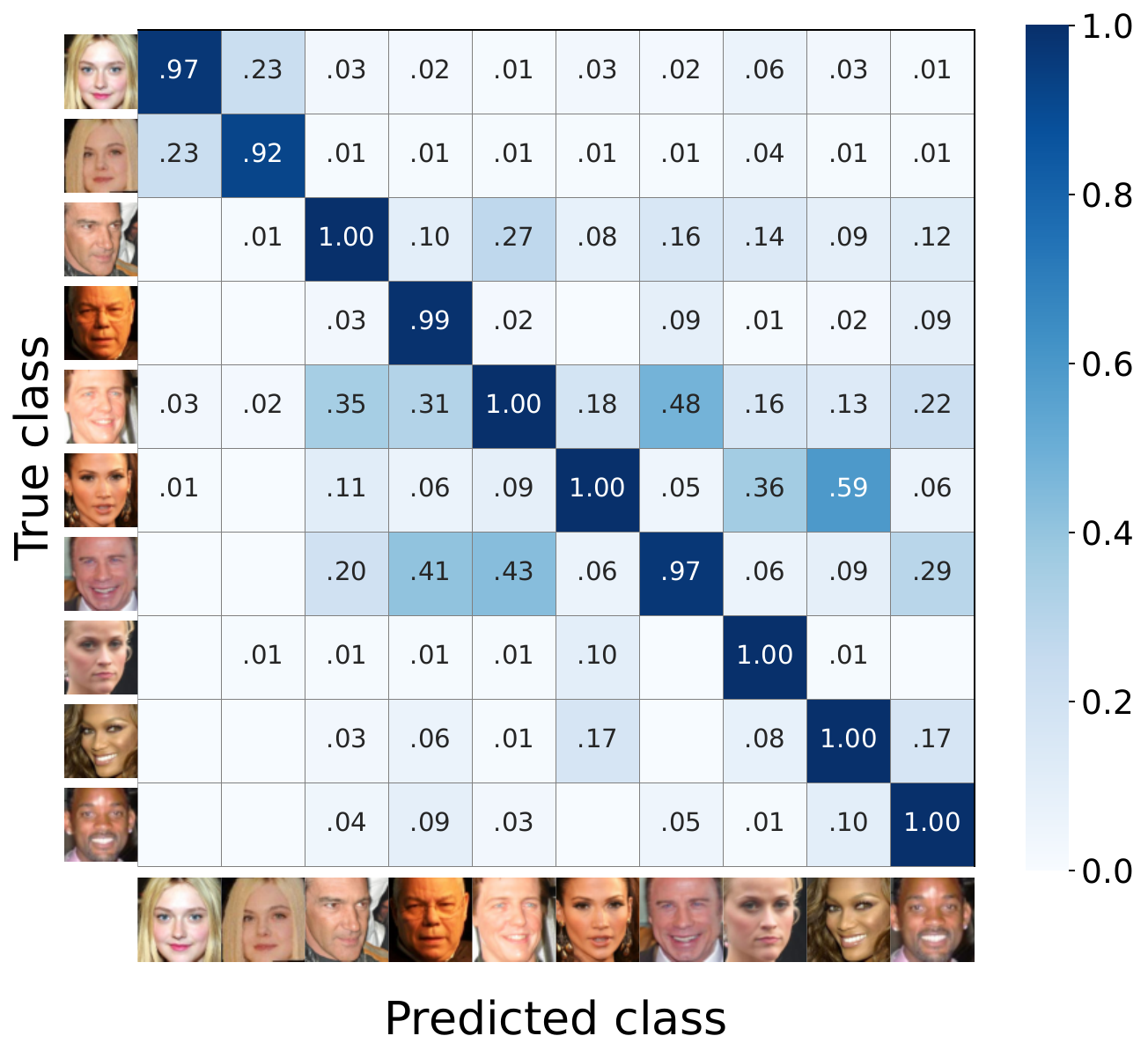}
    \caption{\faal}
  \end{subfigure}
  \hfill
  \begin{subfigure}{0.68\columnwidth}
    \centering
    \includegraphics[width=\linewidth]{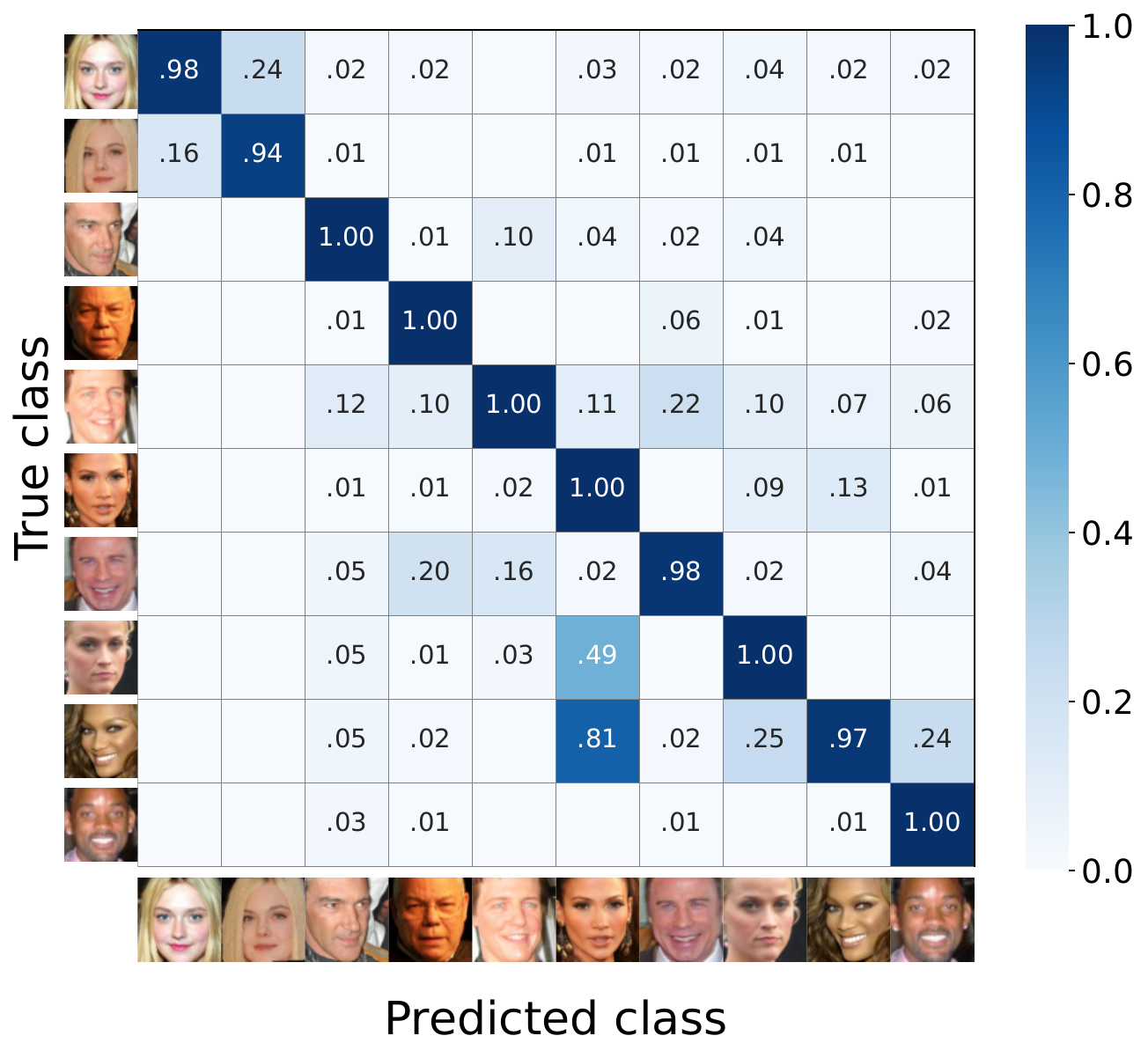}
    \caption{\specnorm}
  \end{subfigure}
  \hfill
  \begin{subfigure}{0.68\columnwidth}
    \centering
    \includegraphics[width=\linewidth]{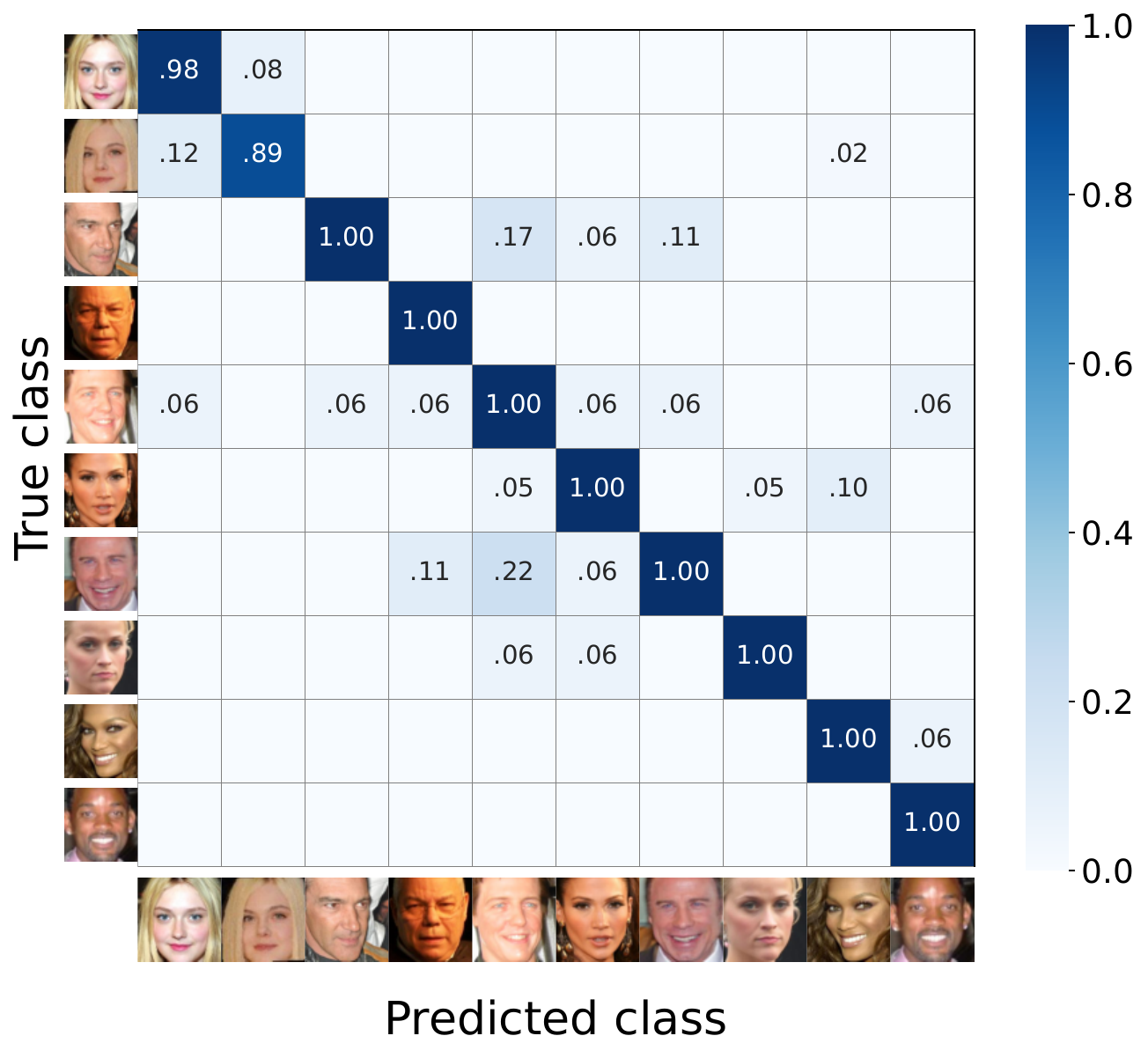}
    \caption{\defense{}}
  \end{subfigure}
    \caption{
    Confusion matrices for different methods on adversarial examples produced with \emph{targeted eyeglass attack} against the
    \pubfigsiblings setup. Diagonals indicate source-class robust accuracy; off-diagonals are misclassifications. To keep the visualization
    representative, we show heatmaps from a single randomly selected run out of the ten repetitions rather than averaged results.}
    \label{fig:pubfig_siblings_tgt_attack_heatmaps}
\end{figure*}
\fi

\section{Stability}
\label{app:stability}

\begin{figure}[H]
    \centering
\includegraphics[width=0.7\linewidth]{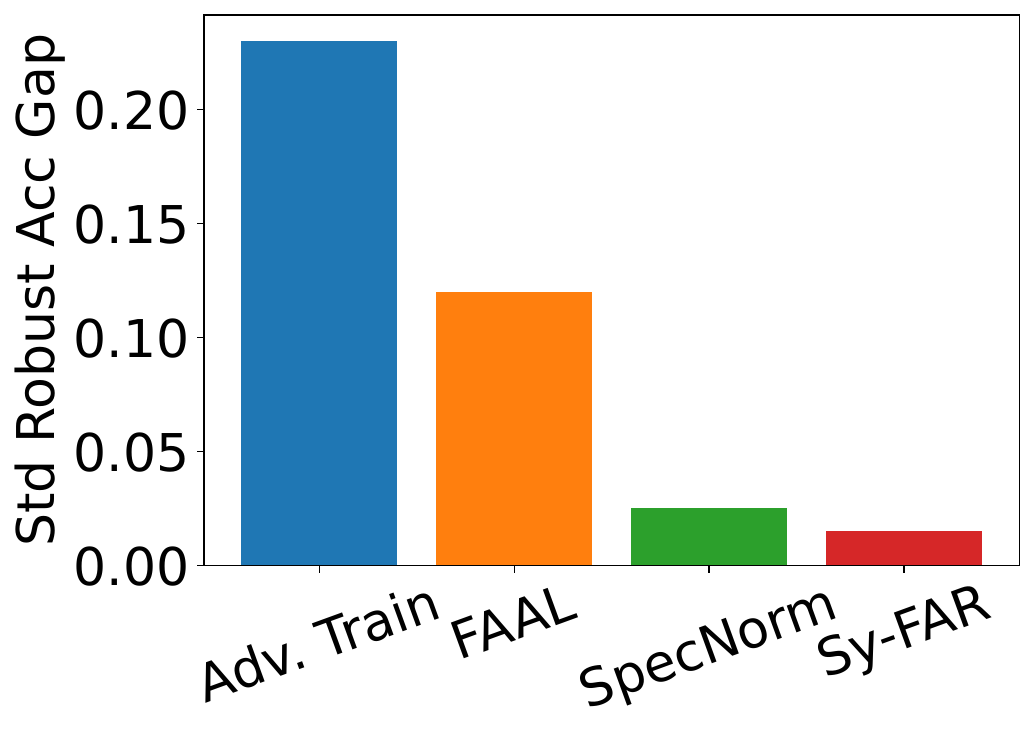}
    \caption{
    Standard deviation in robust accuracy across ten runs in \pubfig under untargeted glass attack. }
    \label{fig:stability-bar1}
\end{figure}

\begin{figure}[H]
    \centering
    \includegraphics[width=0.7\linewidth]{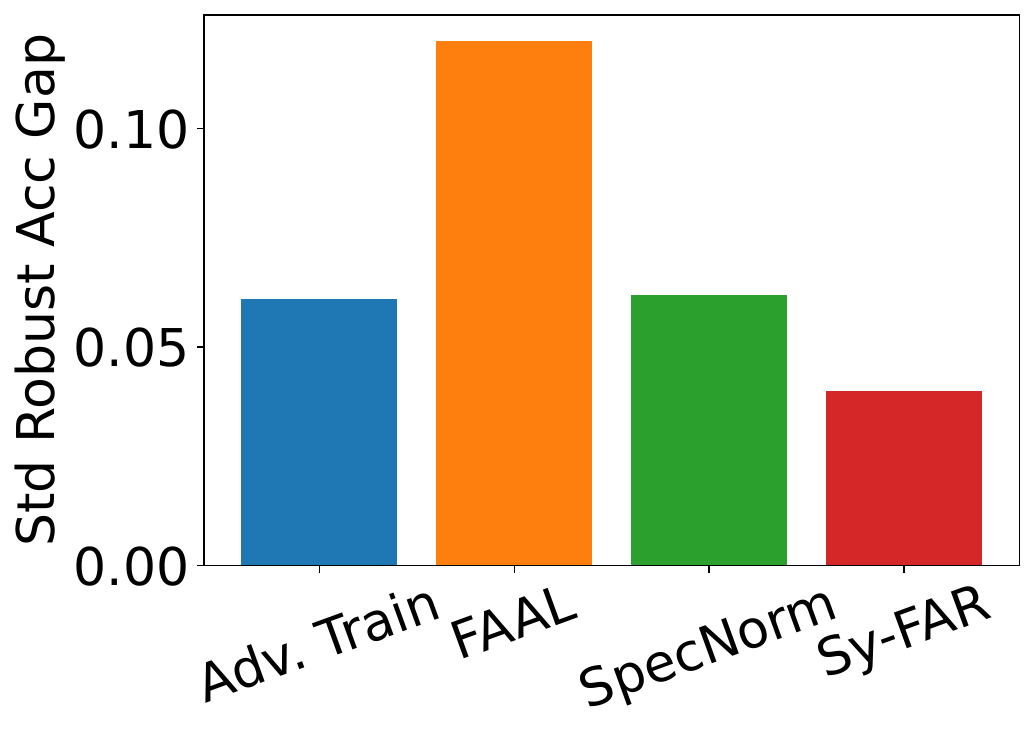}
    \caption{
    Standard deviation in worst-class accuracy across ten runs in \pubfig under untargeted glass attack.} 
    \label{fig:stability-bar2}
\end{figure}

\begin{figure}[H]
    \centering
    \includegraphics[width=0.7\linewidth]{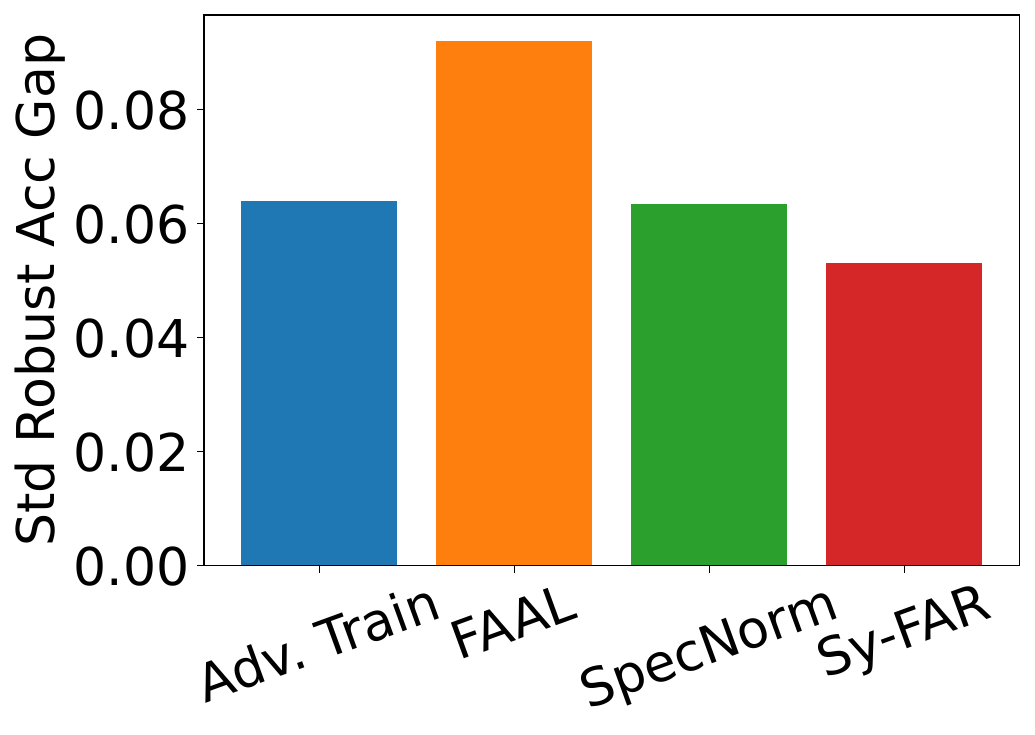}
    \caption{
   Standard deviation in accuracy gap (min-max) across ten runs in \pubfig under untargeted glass attack.}
    \label{fig:stability-bar3}
\end{figure}

Figs.~\ref{fig:stability-bar1}--\ref{fig:stability-bar3} demonstrate \defense{} consistently optimize adversarial robustness and fairness across different runs.

\ifArxiv
\section{Additional Subgroup Fairness Results}
\label{app:subgroup}

Complementing the results from \secref{sec:subgroup:res},
\figref{fig:grouped_color} shows the robust accuracy for ethnic groups (white vs.\ non-white) on \pubfigsiblings, while Tables~\ref{tab:gaps-gender}--\ref{tab:gaps-ethnicity} report the benign and robust accuracy gaps between gender and ethnicity groups across all face-recognition setups we consider.

\begin{table*}[t!]
\centering
\small
\setlength{\tabcolsep}{6pt}
\renewcommand{\arraystretch}{1.1}
\begin{tabular}{@{}l
                c c
                c c
                c c@{}}
\toprule
\multirow{2}{*}{\textbf{Method}}
& \multicolumn{2}{c}{\textbf{\pubfig}}
& \multicolumn{2}{c}{\textbf{\pubfigsiblings}}
& \multicolumn{2}{c}{\textbf{\pubfigvit}} \\
\cmidrule(lr){2-3}\cmidrule(lr){4-5}\cmidrule(lr){6-7}
& \textbf{Benign Gap}~$\downarrow$ & \textbf{Robust Gap}~$\downarrow$
& \textbf{Benign Gap}~$\downarrow$ & \textbf{Robust Gap}~$\downarrow$
& \textbf{Benign Gap}~$\downarrow$ & \textbf{Robust Gap}~$\downarrow$ \\
\midrule
Adv. Train
& 0.044  & 0.0777
& 0.023  & 0.1064
& 0.0323 & 0.0119 \\
FAAL
& 0.014  & 0.1138
& 0.028  & 0.3653
& 0.0523 & 0.0785 \\
SpecNorm
& 0.024  & 0.0378
& 0.015  & 0.3216
& 0.0431 & 0.1070 \\
\textbf{Sy-FAR}
& \textbf{0.012} & \textbf{0.0238}
& \textbf{0.013} & \textbf{0.0433}
& \textbf{0.0429} & \textbf{0.0070} \\
\bottomrule
\end{tabular}
\caption{
\textbf{Benign} and \textbf{robust} accuracy gaps across three setups (\pubfig, \pubfigsiblings, \pubfigvit) for the \emph{gender (male/female) subgroup}. Lower is better ($\downarrow$).}
\label{tab:gaps-gender}
\end{table*}

\begin{figure}[H]
    \centering
    \includegraphics[width=0.45\textwidth]{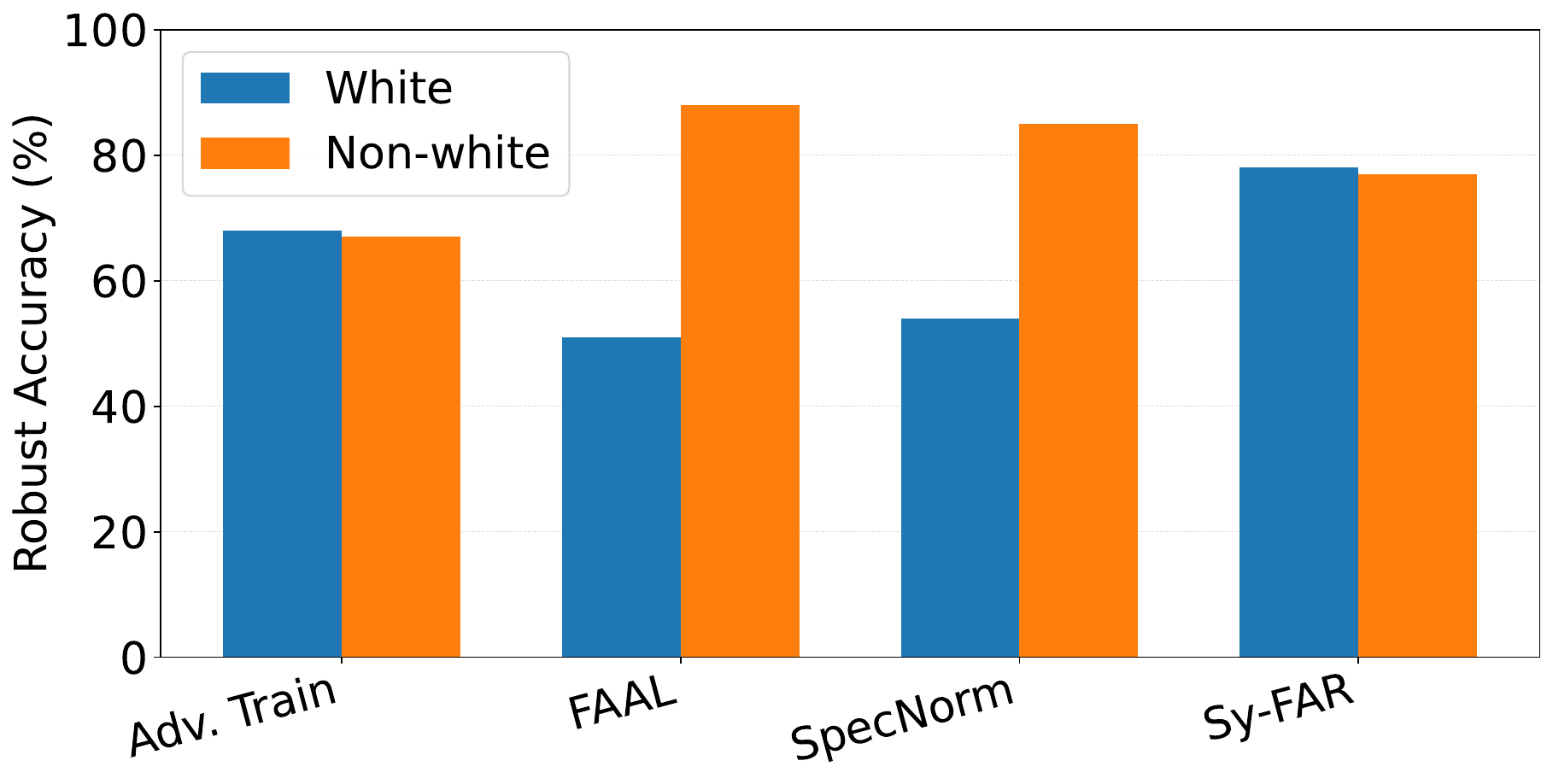}
    \caption{
    Subgroup robust accuracy by ethnicity (white vs.\ non-white) on \pubfigsiblings. 
    }
    \label{fig:grouped_color}
\end{figure}

\begin{table*}[t!]
\centering
\small
\setlength{\tabcolsep}{6pt}
\renewcommand{\arraystretch}{1.1}
\begin{tabular}{@{}l
                c c
                c c
                c c@{}}
\toprule
\multirow{2}{*}{\textbf{Method}}
& \multicolumn{2}{c}{\textbf{\pubfig}}
& \multicolumn{2}{c}{\textbf{\pubfigsiblings}}
& \multicolumn{2}{c}{\textbf{\pubfigvit}} \\
\cmidrule(lr){2-3}\cmidrule(lr){4-5}\cmidrule(lr){6-7}
& \textbf{Benign Gap}~$\downarrow$ & \textbf{Robust Gap}~$\downarrow$
& \textbf{Benign Gap}~$\downarrow$ & \textbf{Robust Gap}~$\downarrow$
& \textbf{Benign Gap}~$\downarrow$ & \textbf{Robust Gap}~$\downarrow$ \\
\midrule
Adv. Train   & 0.024 & 0.0937 & 0.044 & 0.1064 & 0.034 & 0.1413 \\
FAAL         & 0.027 & 0.1004          & 0.018 & 0.3653 & 0.026 & 0.2108 \\
SpecNorm     & 0.024 & 0.1778          & 0.021 & 0.3174 & 0.022 & 0.1220 \\
\textbf{Sy-FAR} 
             & \textbf{0.011} & \textbf{0.0239} & \textbf{0.009} & \textbf{0.1058} & \textbf{0.008} & \textbf{0.1120} \\
\bottomrule
\end{tabular}
\caption{
\textbf{Benign} and \textbf{robust gaps} across three setups (\pubfig, \pubfigsiblings, \pubfigvit) for the \emph{subgroup (ethnicity: White vs. non-White)}. Lower is better ($\downarrow$).}
\label{tab:gaps-ethnicity}
\end{table*}
\fi

\ifArxiv
\section{Face-Recognition Model Training}
\label{app:training}

The training process of the face-recognition models follows prior work~\cite{tong2021facesec}.
We begin with a pre-trained backbone (i.e., feature extractor), to which we attach a randomly initialized fully connected layer followed by a softmax activation.
Subsequently, training is conducted in an end-to-end manner, meaning that all layers of the network are optimized jointly rather than freezing any subset.
To account for stochastic variations due to random initialization, the entire training procedure is repeated ten times, each time starting from a new random seed. This repetition allows us to estimate the stability and consistency of the observed results. 

\fi 

\section{Scalability and Run-Time Analysis}
\label{app:scalability}

To evaluate the computational scalability of Sy-FAR, we analyze the complexity and run-time contribution of the proposed symmetry regularization term across datasets with different numbers of classes. 
During training, nearly all computation time is spent on the forward and backward propagation through the deep network and on adversarial example generation--these steps involve billions of floating-point operations. In contrast, the symmetry penalty performs only lightweight arithmetic operations over class-pair statistics derived from the model’s existing softmax outputs. The penalty does not require additional forward or backward passes, nor does it alter the adversarial optimization process.
Although the number of class pairs grows quadratically with the number of classes ($C(C{-}1)/2$), each symmetry computation is an extremely small GPU tensor operation over already available data. Empirically, we find that:
\begin{itemize}[nosep]
    \item For 10 classes (e.g., PubFig), the symmetry term accounts for $<$0.001\% of the total training time per batch.
    \item For 100 classes (e.g., CIFAR-100), the cost increases slightly to 0.01--0.05\%.
    \item Even for 1,000 classes (e.g., ImageNet-scale settings), the overhead remains $<$0.5\%.
\end{itemize}

In other words, the computation remains negligible compared to the dominant convolutional and adversarial training operations. As the number of classes increases, the total epoch time rises modestly due to the linear growth of the final fully connected (FC) layer and softmax computations, while the symmetry term--despite its theoretical $O(C^2)$ complexity--still contributes a negligible fraction of the total run time.

\ifArxiv
\tabref{tab:scalability} summarizes the empirical estimates. The epoch durations are approximate and provided only to illustrate proportional scaling, not to claim fixed absolute values.

\begin{table*}[!t]
\centering
\small
\setlength{\tabcolsep}{5pt}
\resizebox{\textwidth}{!}{
\begin{tabular}{c c c c c}
\hline
\textbf{\#Classes (C)} & \textbf{Pairs ($\approx C(C{-}1)/2$)} & \textbf{Share of Training Time} & \textbf{Epoch Time (min, $\approx$)} & \textbf{Symmetry Time per Epoch} \\
\hline
10   & 45        & $< 0.001\%$      & $\approx 35$  & $< 0.02$ s \\
100  & 4,950     & $0.01$--$0.05\%$ & $\approx 40$  & $0.2$--$1.2$ s \\
1000 & 499,500   & $0.1$--$0.5\%$   & $\approx 50$  & $3$--$15$ s \\
\hline
\end{tabular}%
}
\caption{Computation cost of the symmetry regularization term as the number of classes increases. The epoch times are approximate and shown to illustrate scalability.}
\label{tab:scalability}
\end{table*}
\fi

\section{Extended Attack Evaluation: Face Masks}
\label{sec:strong_attacks}

To evaluate Sy-FAR under more realistic physical threats, we introduce a \emph{face-mask attack} inspired by FACESEC~\cite{tong2021facesec}. Unlike eyeglass attacks, which apply \emph{pixel-level}, small, localized, high-frequency perturbations around the eyes~\cite{sharif2016accessorize}, the face-mask attack operates at the \emph{grid level}, occluding the lower face and inducing a large, low-frequency perturbation that removes key shape and landmark information. Despite this stronger and fundamentally different threat model, Sy-FAR consistently achieves the best \emph{symmetry loss}, \emph{source- and target-class robustness}, and \emph{fairness} (Tables~\ref{tab:sym_mask_untgtd}--\ref{tab:tgtfair_mask_untgtd}), demonstrating that its symmetry-based regularization remains effective even under
severe facial occlusion.

\begin{table*}[t]
\centering
\small
\setlength{\tabcolsep}{5pt}
\renewcommand{\arraystretch}{1.1}
\begin{tabular}{@{}l c c c c c c@{}}
\toprule
\multirow{2}{*}{\textbf{Method}} 
  & \multicolumn{2}{c}{\textbf{\pubfig}} 
  & \multicolumn{2}{c}{\textbf{\pubfigsiblings}} 
  & \multicolumn{2}{c}{\textbf{\pubfigvit}} \\
\cmidrule(lr){2-3}\cmidrule(lr){4-5}\cmidrule(lr){6-7}
  & \textbf{Asym. Gap}~$\downarrow$ & \textbf{Sym. Loss}~$\downarrow$
  & \textbf{Asym. Gap}~$\downarrow$ & \textbf{Sym. Loss}~$\downarrow$
  & \textbf{Asym. Gap}~$\downarrow$ & \textbf{Sym. Loss}~$\downarrow$ \\
\midrule
Adv. Train & 0.6183 & 2.0856 & 0.4048 & 1.1772 & 0.3602 & 1.4638 \\
\faal       & 0.2117 & 0.7686 & 0.3333 & 1.0288 & 0.3962 & 1.8115 \\
\specnorm   & 0.2475 & 0.8215 & 0.3418 & 1.0426 & 0.3462 & 1.5138 \\
\textbf{Sy-FAR} & \textbf{0.1748} & \textbf{0.4946} & \textbf{0.3117} & \textbf{0.9276} & \textbf{0.2974} & \textbf{1.0920} \\
\bottomrule
\end{tabular}
\caption{Impact of different training methods on symmetry under the untargeted \textit{face-mask attack}. Reported are the Max Asymmetry Gap and Symmetry Loss on \pubfig, \pubfigsiblings, and \pubfigvit using VGG16 and ViT backbones, averaged over ten runs.}
\label{tab:sym_mask_untgtd}
\end{table*}

\begin{table*}[t]
\centering
\small
\setlength{\tabcolsep}{5pt}
\renewcommand{\arraystretch}{1.1}
\begin{tabular}{@{}l c c c c c c c c c c c c@{}}
\toprule
\multirow{2}{*}{\textbf{Method}} 
  & \multicolumn{4}{c}{\textbf{\pubfig}} 
  & \multicolumn{4}{c}{\textbf{\pubfigsiblings}} 
  & \multicolumn{4}{c}{\textbf{\pubfigvit}} \\
\cmidrule(lr){2-5}\cmidrule(lr){6-9}\cmidrule(lr){10-13}
  & \textbf{Clean}~$\uparrow$ & \textbf{Robust}~$\uparrow$ & \textbf{Min}~$\uparrow$ & \textbf{Gap}~$\downarrow$
  & \textbf{Clean}~$\uparrow$ & \textbf{Robust}~$\uparrow$ & \textbf{Min}~$\uparrow$ & \textbf{Gap}~$\downarrow$
  & \textbf{Clean}~$\uparrow$ & \textbf{Robust}~$\uparrow$ & \textbf{Min}~$\uparrow$ & \textbf{Gap}~$\downarrow$ \\
\midrule
Adv. Train & 97.05 & 56.35 & 1.48 & 88.01 & 95.60 & 64.18 & 25.33 & 60.45 & 82.26 & 54.46 & 17.65 & 65.69 \\
\faal       & 98.60 & 79.36 & 57.78 & 39.44 & 93.04 & 68.54 & 26.67 & 58.33 & 81.06 & 57.97 & 20.82 & 69.82 \\
\specnorm   & 98.10 & 77.02 & 55.85 & 38.42 & 95.75 & 72.04 & 28.33 & 60.84 & 82.19 & 56.61 & 26.16 & 63.63 \\
\textbf{Sy-FAR} & \textbf{99.09} & \textbf{84.76} & \textbf{72.88} & \textbf{27.12} & \textbf{95.76} & \textbf{77.05} & \textbf{44.12} & \textbf{55.88} & \textbf{83.87} & \textbf{65.06} & \textbf{33.89} & \textbf{58.96} \\
\bottomrule
\end{tabular}
\caption{Performance comparison across training methods under the untargeted \textit{face-mask attack}. Reported are Clean, Robust, and Minimum (Worst-Class) Accuracies, and Accuracy Gap (Min–Max diagonal difference) on \pubfig, \pubfigsiblings, and \pubfigvit, averaged over ten runs.}
\label{tab:perf_mask_untgtd}
\end{table*}

\begin{table*}[t]
\centering
\small
\setlength{\tabcolsep}{5pt}
\renewcommand{\arraystretch}{1.1}
\begin{tabular}{@{}l c c c c c c c c c@{}}
\toprule
\multirow{2}{*}{\textbf{Method}} 
  & \multicolumn{3}{c}{\textbf{\pubfig}} 
  & \multicolumn{3}{c}{\textbf{\pubfigsiblings}} 
  & \multicolumn{3}{c}{\textbf{\pubfigvit}} \\
\cmidrule(lr){2-4}\cmidrule(lr){5-7}\cmidrule(lr){8-10}
  & \textbf{MinTgt}~$\uparrow$ & \textbf{MaxTgt}~$\downarrow$ & \textbf{Std}~$\downarrow$
  & \textbf{MinTgt}~$\uparrow$ & \textbf{MaxTgt}~$\downarrow$ & \textbf{Std}~$\downarrow$
  & \textbf{MinTgt}~$\uparrow$ & \textbf{MaxTgt}~$\downarrow$ & \textbf{Std}~$\downarrow$ \\
\midrule
Adv. Train & 0.0049 & 0.2736 & 0.0831 & 0.0000 & 0.2485 & 0.0882 & 0.0088 & 0.3719 & 0.0900 \\
\faal       & 0.0038 & 0.2913 & 0.0904 & 0.0000 & 0.3432 & 0.0950 & 0.0016 & 0.3683 & 0.0872 \\
\specnorm   & 0.0078 & 0.2435 & 0.0801 & 0.0044 & 0.2595 & 0.0899 & 0.0055 & 0.2461 & 0.0933 \\
\textbf{Sy-FAR} & \textbf{0.0105} & \textbf{0.1991} & \textbf{0.0623} & \textbf{0.0062} & \textbf{0.2146} & \textbf{0.0669} & \textbf{0.0115} & \textbf{0.1907} & \textbf{0.0582} \\
\bottomrule
\end{tabular}
\caption{Target-class fairness under the untargeted \textit{face-mask attack}. Reported are the minimum and maximum misclassification shares (MinTgt and MaxTgt) and their standard deviation across classes, for \pubfig, \pubfigsiblings, and \pubfigvit setups, averaged over ten runs.}
\label{tab:tgtfair_mask_untgtd}
\end{table*}

\end{document}